\documentclass[twoside]{article}
\usepackage{ecj,palatino,epsfig,latexsym,natbib}

\usepackage{xcolor}
\usepackage{amsmath}
\usepackage{amssymb}

\usepackage{graphicx}

\usepackage{colortbl} 
\usepackage{soul}
\usepackage[figuresright]{rotating}

\usepackage{algorithm}
\usepackage{algpseudocode}
\usepackage{url}
\usepackage{blkarray, bigstrut}
\usepackage{subcaption}
\usepackage{afterpage}
\usepackage{lscape}

\usepackage{listings}
\usepackage{verbatim}

\algrenewcommand\algorithmicrequire{\textbf{Data:}}
\algrenewcommand\algorithmicensure{\textbf{Result:}}

\newcommand\topstrut[1][1.2ex]{\setlength\bigstrutjot{#1}{\bigstrut[t]}}
\newcommand\botstrut[1][0.9ex]{\setlength\bigstrutjot{#1}{\bigstrut[b]}}

\usepackage[symbol]{footmisc}

\begin{document}

\ecjHeader{x}{x}{xxx-xxx}{2022}{Informed Down-Sampled Lexicase Selection}{R. Boldi, M. Briesch, D. Sobania, A. Lalejini, T. Helmuth, F. Rothlauf, C. Ofria and L. Spector}
\title{Informed Down-Sampled Lexicase Selection: Identifying productive training cases for efficient problem solving}
\author{\name{\vspace{-2em}\bf Ryan Boldi\thanks{\;\;Both authors contributed equally.} } 
        \hfill         \addr{rbahlousbold@umass.edu}\\ 
        \addr{
        University of Massachusetts, Amherst, MA 01003, USA}
\AND
       \name{\bf Martin Briesch}\footnotemark[1]
        \hfill \addr{briesch@uni-mainz.de}\\
        \addr{
        Johannes Gutenberg University, 
        Mainz, 55128, Germany}
\AND
       \name{\bf Dominik Sobania} \hfill \addr{dsobania@uni-mainz.de}\\
        \addr{
        Johannes Gutenberg University, 
        Mainz, 55128, Germany}
\AND
       \name{\bf Alexander Lalejini} \hfill \addr{lalejina@gvsu.edu}\\
        \addr{
        Grand Valley State University, Allendale, MI 49401, USA
}
\AND
       \name{\bf Thomas Helmuth} \hfill \addr{thelmuth@hamilton.edu}\\
        \addr{
        Hamilton College, Clinton, NY, 13323, USA}
\AND
       \name{\bf Franz Rothlauf} \hfill \addr{rothlauf@uni-mainz.de}\\
        \addr{
        Johannes Gutenberg University, 
        Mainz, 55128, Germany}
\AND
    \name{\bf Charles Ofria} \hfill \addr{ofria@msu.edu}\\
        \addr{
        Michigan State University, 
        East Lansing, MI 48824, USA}
\AND
       \name{\bf Lee Spector} \hfill \addr{lspector@amherst.edu}\\
        \addr{
        Amherst College, Amherst, MA 01002, USA} \\
     }

\maketitle
\begin{abstract}


Genetic Programming (GP) often uses large training sets and requires all individuals to be evaluated on all training cases during selection. Random down-sampled lexicase selection evaluates individuals on only a random subset of the training cases allowing for more individuals to be explored with the same amount of program executions.
However, sampling randomly can exclude important cases from the down-sample for a number of generations, while cases that measure the same behavior (synonymous cases) may be overused.
In this work, we introduce Informed Down-Sampled Lexicase Selection. This method leverages population statistics to build down-samples that contain more distinct and therefore informative training cases.
Through an empirical investigation across two different GP systems (PushGP and Grammar-Guided GP), we find that informed down-sampling significantly outperforms random down-sampling on a set of contemporary program synthesis benchmark problems. Through an analysis of the created down-samples, we find that important training cases are included in the down-sample consistently across independent evolutionary runs and systems. We hypothesize that this improvement can be attributed to the ability of Informed Down-Sampled Lexicase Selection to maintain more specialist individuals over the course of evolution, while still benefiting from reduced per-evaluation costs.

\end{abstract}


\section{Introduction}
 
In Evolutionary Computation, we often use large sets of training data to evaluate the quality of candidate solutions. 
For instance, most Genetic Programming (GP) systems evaluate programs using input/output examples (training cases) that specify the expected behavior of a correct program. 
Many GP selection strategies aggregate each program's performance across all training cases to produce one \emph{fitness} score that can be used for selection. 
In contrast, lexicase selection \citep{Spector2012AssessmentOP, Helmuth:2015:ieeeTEC} avoids aggregation and considers each training case separately, which has been shown to improve diversity maintenance \citep{Helmuth16Diversity, dolson_ecological_2018} and problem-solving success across a wide range of domains \citep{moore2017lex, Metevier2019, aenuguLCS2019, Ding2022optimizing, lalejini_artificial_2022}. 

However, standard lexicase selection requires that we evaluate all individuals on all training cases, which can be computationally expensive when evaluation is non-trivial.  
To reduce lexicase selection's computational cost, recent work introduced down-sampled lexicase selection \citep{moore2017lex, Hernandez:2019:subsampling, Ferguson2020Characterizing}.
In down-sampled lexicase selection, the training set is randomly down-sampled, reducing the number of training case evaluations required to assess the quality of each candidate solution.
This in turn reduces the cost of evaluating an entire set of individuals, allowing us to reallocate computational resources to other aspects of an evolutionary search (e.g., increasing search time or population size), which can substantially improve problem-solving success \citep{Helmuth2020explaining, Helmuth2021benefits, Hernandez:2019:subsampling}.
However, a naive random down-sample can leave out potentially important training cases, resulting in a loss of diversity 
\citep{Ferguson2020Characterizing, helmuth_importance_2020, hernandez_exploration_2022}.

In order to put more computational effort towards evaluating individuals on important training cases, 
we propose \textit{informed down-sampling} (IDS), which uses runtime population statistics to build a down-sample that contains more distinct cases. 
Given a set of solutions, two training cases are distinct from each other if the subsets of solutions that solve each of the two test cases have little-to-no overlap. 
Two training cases are synonymous if the opposite is true: there is substantial overlap between the subsets of solutions that solve each case\footnote{Synonymous cases can also be thought of as cases that have different inputs and outputs yet measure a very similar functionality such that there is a high correlation between individuals' performance on these cases.}. Consequently, Informed down-sampling favors the distinct training cases over synonymous cases when building a down-sample to use for selection. 
We expect these informed down-samples to better maintain unique individuals, increasing overall population diversity while also putting more selection pressure on individuals whose descendants are more likely to solve the problem.  
These unique individuals are often viewed as the stepping-stones for evolution to use in finding a perfect solution program \citep{helmuth_importance_2020}. 

To assess the performance of Informed Down-Sampled Lexicase Selection, we compare lexicase selection without down-sampling (standard lexicase), with random down-sampling, and with informed down-sampling across eight problems from the first and second program synthesis benchmark suites \citep{Helmuth2015psb1, helmuth2021psb2}. We conduct our experiments in two independent GP frameworks, Grammar-Guided Genetic Programming (G3P) \citep{whigham1995grammatically, forstenlechner2016grammar, forstenlechner2017grammar} and PushGP \citep{spector2002autoconstructive, SpectorPush3}.

We find that building a down-sample based on information we collect from the population is a valuable way to improve the success rates of evolutionary runs at a fixed computational cost. Furthermore, simply tracking which cases are distinct, and ensuring they are placed in a down-sample, can significantly improve problem solving performance. Our results provide evidence that informed down-sampling improves the success rate of search in the two GP systems used. By analyzing the composition of down-samples, we also verify that informed down-sampling builds down-samples that contain more informative test cases (e.g., edge cases) than random down-sampling.

\section{Related Work}\label{sec:relatedWork}

In most GP applications, parent selection uses the performance of candidate solutions on a set of training cases to pick individuals that contribute genetic material to the next generation. Most selection algorithms aggregate the scores on these training cases to get a single score per candidate and then select the most fit candidates using tournament selection \citep{brindle_1980}, implicit fitness sharing \citep{Smith93IFS}, fitness proportionate selection \citep{Holland1975AdaptationIN}, or another selection strategy. The fitness aggregation procedure for these methods often results in a loss of semantic information about which training cases the individual performs well on \citep{Krawiec15}, motivating the development of selection strategies that consider each individual's performance on all training cases encountered \citep{VanneschiSemantic2014,goings_ecology_based_2012,deb_fast_2002,horn_niched_1994}.

In contrast, lexicase selection does not aggregate fitness or performance measures \citep{Spector2012AssessmentOP}. 
For each parent selection event, the lexicase selection procedure first places all individuals in the population into a ``parent pool'' (i.e., the pool of individuals eligible to be selected). 
To select a parent, lexicase selection shuffles the training cases into a random ordering, and each training case is considered in sequence.
For each training case, the parent pool is filtered down to just the individuals that have the best (or tie for the best) performance, removing all but the best candidates from further consideration. 
If there is only one individual that remains in the pool during this filtering process, this individual is selected. If the training cases are exhausted and there are still individuals in the pool, one of these individuals is selected at random.

Meanwhile, many variants of lexicase selection have been proposed for use in different problems or domains. For example, epsilon lexicase selection \citep{lacavaEpLex, moore2017lex}, batch lexicase selection \citep{aenuguLCS2019, sobania_2022_batch}, gradient lexicase selection \citep{Ding2022optimizing}, lexicase selection for GAs \citep{Metevier2019}, weighted shuffle lexicase selection \citep{Troise:2017:GPTP}, and fast lexicase selection \citep{ding2022scale}.

One of the most promising variants of lexicase selection is down-sampled lexicase selection, which was first proposed for expensive evolutionary robotics runs by \cite{moore2017lex} and later formalized by \cite{Hernandez:2019:subsampling} for GP runs. So far, down-sampled lexicase selection increased the success and generalization rates for a variety of problems \citep{Ferguson2020Characterizing}. Down-sampled lexicase selection works by randomly sampling once in each generation the training set to create a smaller set of cases. These cases are then used to perform all selection events in the population for that one generation. This limitation on the number of test cases reduces the computational costs of evaluating the individuals, which is usually one of the most expensive operations in evolutionary runs. These savings could be used to perform computationally cheaper GP runs, increase the population size, or run evolution for more generations. 

Down-sampled lexicase selection has also been found to significantly outperform regular lexicase selection in a variety of program synthesis benchmarks \citep{Hernandez:2019:subsampling, Ferguson2020Characterizing, Helmuth2020explaining, Helmuth2021benefits,helmuth_benchmarking_2020}.  
However, creating a down-sample randomly can exclude important training cases from the current down-sample for a number of generations \citep{hernandez_exploration_2022}, while synonymous cases may be overused. 
As a first attempt at changing the composition of cases in the down-sample, \cite{Boldi_2022_Environmental} explored using a rolling down-sample and a disjoint down-sample for lexicase selection runs. While the results were neutral-if-not-negative, they highlighted the presence of synonymous cases in practice and suggest that an attempt at mediating the time put into evaluating individuals on these synonymous cases might improve search performance. 

Other work has used coevolutionary methods to identify a smaller, representative set of training cases that can be used to assess candidate solutions instead of using the entire training set ~\citep{Schmidt2005coevolving,Schmidt2008CoevolutionOF, sekanina2012CGP}. 
Our proposed informed down-sampling method also results in a compressed training set that is roughly as informative as the set of all available data but does not require an additional coevolutionary process, making it more easily compatible with existing GP systems.    
Another important example is the use of random down-sampling to improve performance of AutoML runs that use GP to evolve machine learning pipelines \citep{zogaj_doing_2021}. This work, however, did not include any form of non-random down-sampling such as informed down-sampling.

In the broader machine learning community, random down-sampling is used to generate mini-batches for stochastic gradient descent \citep{ruder_overview_2017}, and many forms of non-random down-sampling are used to detect hard or informative parts of the training data \citep{Loshchilov2015OnlineBS,paul2021deep, Chrysakis2020OnlineCL}. 
Many of these machine learning methods seek to identify a `coresets', which is a small summarization of the training data such that solving the task on a coreset can provably yield solution with tightly bounded error on the complete training set \citep{Bachem2017PracticalCC,jubran_introduction_2019}.    
Indeed, many of these more systematic approaches to reducing training set size have yet to be explored in the context of GP, especially in combination with the lexicase parent selection method. 

\section{Informed Down-Sampling}\label{sec:IDS}

Random down-sampling in lexicase selection, while effective in reducing computational costs, can inadvertently dilute the selection pressure by picking synonymous cases, thereby concealing the true quality of candidate solutions. Worse still, important cases might be overlooked due to stochasticity, leading to important genetic material being lost to the population. Overcoming these shortcomings is of paramount importance, especially considering the increasingly complex datasets and challenges faced in GP applications.
Informed down-sampling offers a promising solution. It intelligently selects more diverse and representative training cases based on runtime statistics, enhancing the ability of lexicase selection to accurately assess solution quality, preserve useful genetic information, and ultimately yield more robust and higher-quality solutions.


We suggest two methods of building an informed down-sample. 
First, we explore the idealized effectiveness of informed down-sampling by presenting it with full information. 
This method requires evaluating the entire population on all training cases, performing the same number of program executions per generation as normal lexicase selection. Therefore, informed down-sampling with full information cannot capitalize on the computational savings afforded by random down-sampling. 
However, the full information approach provides useful intuition for building an informed down-sample, allowing us to measure the problem-solving success of our sampling approach under idealized conditions. 

Next, we present an approach for creating an informed down-sample that reduces the number of per-generation evaluations required for selection (relative to standard lexicase selection). 
This second approach, referred to as the ``sparse information'' approach, estimates the distinctness of training cases based on a sample of individuals from the parent population.
Indeed, building an informed down-sample using sparse information results in nearly the same per-generation evaluation savings as when using random down-sampling.

\subsection{Building an Informed Down-Sample with Full Information}
\label{sec:ids:full-info}

In our informed down-sampling approach with full information, we create one down-sample of training cases per generation. Then, to select a parent, we use candidate solution performances on only the \emph{sampled} training cases. To construct an informed down-sample with full information, we evaluate all members of the population on all training cases. 
In this work, training cases are evaluated on a pass/fail basis. 
Next, we construct the ``solve vector'' $S_{j}$ for each training case $c_j$, which is a vector of binary values that specifies which individuals in the population have solved the training case $c_j$.

Figure~\ref{fig:distanceVisual} provides an example set of binary solve vectors for a set of five training cases and a population of six individuals. The columns in this matrix $I_i$ describe the performance of the $i^{\text{th}}$ individual on all cases. A value of 1 at $(I_i, c_j)$ implies that the $i^\text{th}$ individual solved the $j^\text{th}$ training case ($\textrm{error}=0$), or $S_{j}^i = 1$. 
The number of columns corresponds to the population size.


\begin{figure}
    \centering
    $$
    \begin{blockarray}{ccccccc}
    & I_1 & I_2 & I_3 & I_4 & I_5 & I_6 \\
    \begin{block}{c[cccccc]} S_{1} &0 & 1 & 0 & 1 & 1 & 0\topstrut\\
    S_2 & 1 & 1 & 0 & 0 & 1 & 1\\
    S_3 & 1 & 0 & 1 & 1 & 0 & 1\\
    S_4 & 0 & 1 & 0 & 0 & 1 & 1 \\
    S_5 & 0 & 1 & 0 & 1 & 1 & 0\botstrut\\
    \end{block}
    \end{blockarray}
    $$

    \caption{Example of the data structure that is used to determine distances between cases. $c_{1,\dots,5}$ are cases, with their respective solve vectors $S_{1,\dots, 5}$, and $I_{1,\dots,6}$ are individuals. The entry at $S_j$ and $I_i$ represents whether the $i^\text{th}$ individual solved the $j^\text{th}$ training case or not. The binary solve vectors $S_j$ can be read off as the respective row for the $j^\text{th}$ case. The distance between two cases, $D(c_x, c_y)$, is the Hamming distance between their respective solve vectors (the rows for each case). For example, $D(c_1, c_2) = 3$ and $D(c_2, c_3) = 4$.}
    \label{fig:distanceVisual}
\end{figure}

We define the distance between two training cases $D(c_x, c_y):=\text{Hamming}(S_{x}, S_{y})$ where $\text{Hamming}(\mathord{\cdot} , \mathord{\cdot})$ is the Hamming distance between two vectors. For binary vectors, the distance function is defined as: $D(c_x, c_y) = \sum_{i=1}^{p}|S_{x}^i - S_{y}^i|$. Thus, two training cases that are solved by the  same set of individuals are deemed to have $D(c_1, c_2) = 0$ and are  called ``synonymous cases". For example, for the cases in Figure~\ref{fig:distanceVisual}, $c_1$ and $c_5$ have identical solve vectors, and therefore are synonymous ($D(c_1, c_5) = 0$).

We think of this distance function as indicating the joint information contained in a pair of cases. Two cases that have exactly the same individuals solving them (i.e. are synonymous) have little to no joint information because having both of the cases in the sample would be about as informative as just having one of them. Two cases that have a high distance from each other, due to being solved by different subsets of the population, have high joint information as each case is responsible for informing the system about the performance of one set of individuals. Having both of these cases, as opposed to one alone, would be a more faithful approximation of using the full training set.

We use an algorithm based on Farthest First Traversal  \citep{Hochbaum1985ABP}  to select a down-sample based on pairwise case distances. Our Farthest First Traversal algorithm, is shown in algorithm~\ref{alg:FFT}. Starting with an empty down-sample and a set of cases $\mathcal{T}$, we first add a random case to the down-sample (line 4), and then iteratively add the cases that are maximally far from the closest case to it (5-9). If there are multiple cases with the same maximum minimum distance, ties are broken randomly. The $MinDist_i$ value stores the distance from a given case, $c_i$ to the closest case to it in the down-sample. The $\mathcal{T}.popMaxMinDistCase()$ function removes and returns the case in $\mathcal{T}$ that has the maximum value for $MinDist_i$. Note here that it is often the case that the minimum distances all go to zero at a point during the down-sample formation. At this point, every case left over in the training set has a synonymous case in the down-sample already. When this happens, the farthest first procedure will automatically select cases at random from the training set to fill up the required down-sample size. Figure~\ref{fig:runningExample} shows an example of performing informed down-sampling with full information using the case solve vectors from Figure~\ref{fig:distanceVisual}.

\begin{algorithm}
\caption{Farthest First Traversal Down-Sample Selection}\label{alg:FFT}
\begin{algorithmic}[1]
\Require ${D(\cdot,\cdot)}: D(c_i, c_j) = D(c_j, c_i) = \text{distance from case } i \text{ to case } j, \hfill$ 
$r = \text{ down-sample rate}$
\State $\mathcal{T} \gets $ set of all cases in training set
\State $\mathbf{ds} \gets $ empty set \Comment{the down-sample}
\State $\mathbf{size} \gets r\times |\mathcal{T}|$ \Comment{desired size of down-sample}
\State $\mathbf{ds}.add(\mathcal{T}.popRandomCase())$
\While{$\lVert\mathbf{ds}\rVert < \mathbf{size}$}
\For{every case $c$ in $\mathcal{T}$}
    \State {$MinDist_i \gets $ minimum distance from $c_i$ to any case in $\mathbf{ds}$}
\EndFor
\State $\mathbf{ds}.add(\mathcal{T}.popMaxMinDistCase())$
\EndWhile
\State \bf{return} $\mathbf{ds}$
\end{algorithmic}
\end{algorithm}

\begin{figure}
    \centering
    \begin{subfigure}[b]{0.45\textwidth}
    $$
    D = \begin{blockarray}{cccccc}
    & c_1 & c_2 & c_3 & c_4 & c_5\\
    \begin{block}{c[ccccc]} c_{1} & 0 & 3 & 5 & 2 & 0\topstrut\\
    c_2 & 3 & 0 & 4 & 1 & 3\\
    c_3 & 5 & 4 & 0 & 5 & 5\\
    c_4 & 2 & 1 & 5 & 0 & 2\\
    c_5 & 0 & 3 & 5 & 2 & 0\botstrut\\
    \end{block}
    \end{blockarray}
    $$
    \end{subfigure}
    \begin{subfigure}[b]{0.45\textwidth}
    \vspace{3em}
    $$
    \overbrace{\textbf{ds}=\{c_1\}}^{\text{Random}} \ \ \ \ \rightarrow \overbrace{\textbf{ds}=\{c_1, c_3\}}^{c_3\text{ had max. distance to } c_1}$$
    $$
    \rightarrow \overbrace{\textbf{ds}=\{c_1, c_3, c_2\}}^{c_2\text{ had max. min. distance to $\{c_1, c_3\}$}}
    $$
    \vspace{-3em}
    \end{subfigure}
    \caption{Example running procedure of informed down-sampling with full information to pick a down-sample of size 3 (or $r = \frac{3}{5})$. We have a tabular representation of the distance function $D$ generated by computing the Hamming distance between each pair of cases' solve vectors. Beginning with a randomly selected case $c_1$, we sequentially add the cases that are at the maximum distance to their closest case in the down-sample. The first step is simply finding the case ($c_3$) in the training set with the maximum distance to $c_1$. To select the next case, we need to find, for $c_2$, $c_4$ and $c_5$, which of $c_1$ and $c_3$ is closest to them, respectively, and then which of \emph{those} cases is farthest away. In this example, $c_2$ was added as it had a higher distance (3) to its closest case than did $c_4$ or $c_5$ (2 and 0, respectively). Notice that the cases that were left out, $c_4$ and $c_5$, are synonymous or nearly synonymous with cases already in the down-sample: $c_2$ and $c_1$, respectively.}\label{fig:runningExample}
\end{figure}

\subsection{Building an Informed Down-Sample with Sparse Information}
\label{spareseInfoSection}

Down-sampled lexicase selection's problem-solving benefits stem from the computational savings gained by not evaluating the entire population on the whole training set every generation.
For a fixed computational budget, down-sampling allows more computational resources to be allocated to other aspects of evolutionary search, such as running for more generations or increasing population size.  
As a result, a larger portion of the search space can be explored \citep{Helmuth2021benefits}. 
Informed down-sampling with full information requires the evaluation of all individuals on all training cases in order to construct the down-sample to use in selection. This entire process is counter productive, as we could have just used the initial population evaluation to select individuals and circumvent the entire down-sampling process. The benefit of down-sampling comes from its ability to use sparse information in the individual selection process. Since our aim is to improve on random down-sampling, we must reduce the number of necessary program executions in order to calculate distances between training cases, so that we can benefit from sparse evaluations in both our individual selections and our down-sample creation.

We present two methods to decrease the number of evaluations required for the pairwise distance calculation procedure. The first method, \textit{parent sampling}, samples a proportion $\rho$ of the parents to evaluate the distances for every generation. These parent-samples are evaluated on the \emph{entire} training set. In our runs with a population size of $1000$, if we were to randomly sample $0.01$ (or $\rho=0.01$) of these parents to become the parent sample, these $10$ parents would be evaluated on \emph{all} training cases. This results in case solve vectors of length $10$ that are used to calculate the distances between cases. Distances between cases are determined purely based on these parent-sample evaluations.
We use the distance matrix generated from these parents to estimate the joint informativeness. 

The second method, \textit{scheduled case distance computation}, involves recomputing the distance matrix from the current population every $k$ generations, as opposed to every generation. This schedule reduces the amount of computation required for the evaluation of case distances even further by not performing it every generation. While such a schedule does not update the distances between cases as often, we still re-sample the down-sample based on these distances \emph{every generation}. Due to the stochastic nature of the down-sample selection process (specifically the random selection of the first case), it is likely that the same down-sample will not be used to evaluate the population in consecutive generations.

Both parent sampling and scheduled case distance computation allow us to select a down-sample using less information about individuals while losing only a small amount of information about cases and their similarity.
Algorithm~\ref{alg:IDSTotal} shows how a single generation of parent selection events may be performed using informed down-sampling with sparse information.  

\begin{algorithm}
\caption{Informed Down-Sampling with Sparse Information}
\label{alg:IDSTotal}
\begin{algorithmic}[1]
\Require
\Statex $\mathcal{P}:$ population,
\Statex $\mathcal{T}$:  set of all training cases,
\Statex$k: $ scheduled case distance computation parameter,
\Statex $\rho: $ parent sampling rate, 
\Statex $\mathcal{G}:$  current generation counter,
\Statex $\mathcal{D}: $ case distance matrix. \Comment{all distances initialized to be maximally far}
\Ensure A list of selected parents
\If{$\mathcal{G} \% k == 0$}
\State{$\hat{\mathcal{P}} \gets$ sample $\rho{\times}|\mathcal{P}|$ parents from $\mathcal{P}$}
\State{evaluate $\hat{\mathcal{P}}$ on cases in $\mathcal{T}$}
\State{calculate $\mathcal{D}$ from case solve vectors from solutions in $\hat{\mathcal{P}}$ on cases in $\mathcal{T}$}
\EndIf
\State{$D(\cdot, \cdot) \gets $ distance function derived from indexing into $\mathcal{D}$}
\State{$\textbf{ds} \gets $ create downsample using farthest first traversal down-sampling (See Algo~\ref{alg:FFT})}
\State{$\mathcal{P} \gets$ select $|\mathcal{P}|$ new parents using lexicase selection from $\mathcal{P}$ using $\textbf{ds}$ as cases}
\State \bf{return} $\mathcal{P}$
\end{algorithmic}
\end{algorithm}

\section{Experimental Methods}\label{sec:experiments}

We conducted a series of experiments to study the efficacy of informed down-sampled lexicase selection.  
We compared the performance of informed down-sampled, random down-sampled, and standard lexicase selection on a series of program synthesis benchmark problems. We replicated all experiments in two independent GP systems to test whether our findings are robust across different program representations: PushGP and Grammar Guided Genetic Programming (G3P).


\subsection{Program Synthesis Benchmark Problems}

Our experiments used eight program synthesis benchmark problems from the first and second general program synthesis benchmark suites \citep{Helmuth2015psb1, helmuth2021psb2}: Count Odds, Find Pair, Fizz Buzz, Fuel Cost, Greatest Common Denominator (GCD), Grade, Scrabble Score, and Small or Large. 
These benchmark suites include a variety of introductory programming problems that require the manipulation of multiple data types with complex looping or conditional structures. These benchmark problems are recent, and widely accepted across a variety of GP papers \cite{helmuth_applying_2022}.
The eight problems that we chose to use are well-studied and are commonly used to compare parent selection algorithms in a GP context \citep{sobania2022comprehensive, sobania2022copilot}. 

Each benchmark problem is defined by a set of input/output examples (referred to as \emph{cases}) that specify the desired behavior of a correct program.
For each problem, we split the input/output examples into a training set and a testing set.
During evolution, we assessed program quality using only the training set.
We used the testing set to measure how well a program generalized on examples unseen during evolution.
We consider each input/output example on a pass/fail basis; that is, a program passes a training case if it produces the correct output when run with the associated input. 
A program is a \textit{solution} if it passes all of the training cases; it \textit{generalizes} if it passes all training \emph{and} all testing cases. We refer to runs as ``success" if they result in the production of a generalizing solution.
We used the same training and testing data sets across both PushGP and G3P for each problem to ensure the data available is not biasing  performance.

We include the problem descriptions for each of the eight chosen benchmark problems in Appendix~\ref{appendix:problems}, including their input and output types (Table~\ref{tab:problems}).   
We selected these particular problems to allow us to test informed down-sampling on a set of easy, medium, and hard problems as established by published success rates using PushGP and random down-sampled lexicase selection \citep{Helmuth2021benefits, helmuth_applying_2022}. 

\subsection{Genetic Programming Systems}

PushGP is a system that evolves computer programs in the \emph{Push} programming language, a stack-based language specifically invented for use in genetic programming \citep{spector2002autoconstructive, SpectorPush3}. Push literals are pushed onto one of a set of datatype specific stacks while instructions are also stored on a stack during interpretation. These instructions usually act on data from the stacks and leave their return value on the stacks. Instructions take values from and return results to the appropriately typed stack, including from and to the instruction stack, allowing for programs to use multiple data types and complex conditional execution paradigms. 
In this work, we used the propeller implementation of PushGP. 

G3P uses a context-free grammar in Backus-Naur form to evolve individuals in a desired programming language and supports the use of different data types and control structures \citep{whigham1995grammatically, forstenlechner2016grammar, forstenlechner2017grammar}. To prevent the generation of many invalid solutions during search, we use a tree-based representation instead of the common genotype-phenotype mapping known from classical grammatical evolution \citep{ryan1998grammatical,sobania2020challenges}. 
Our implementation of G3P uses the PonyGE2 framework~\citep{fenton2017ponyge2}.
The code for both GP systems, as well as lists of instructions and grammars used for our experimentation can be found in our web-based supplemental material \citep{boldi2023SupplementIDS}.

Using two independent GP system allows us to better establish whether our findings are system-specific or more broadly applicable beyond a single GP representation.  
PushGP and G3P have many fundamental representational differences that influence how they drive populations through the search space.
By replicating our experiments with both PushGP and G3P, we are better able to evaluate whether any observed performance differences among selection methods are representation-specific. 

\begin{table}[t]
    \centering
    \caption{General and System-Specific Evolution Parameters}
    \begin{tabular*}{\textwidth}{l@{\extracolsep{\fill}}r}
    \hline
    \textbf{General Parameter}     & \textbf{Value} \\
    \hline
    runs per problem & 100 \\
    population size & 1,000 \\
    size of training set  & 200 \\
    size of test set & 1,000 \\
    program execution limit & 60 million \\
    maximum number (base) of generations & 300 \\
    \hline
    \textbf{PushGP Parameter}     & \textbf{Value}  \\
    \hline
    variation operator & UMAD   \\
    UMAD rate & 0.1  \vspace{0.3em}\\
    step limit & 2000 \\
    max initial plushy size & 250 \\
    \hline 
    \textbf{G3P Parameter}     & \textbf{Value}\\
    \hline
    crossover operator & subtree crossover\\
    crossover probability & 0.95\\
    mutation operator & subtree mutation \\
    mutation steps & 1 \\
    maximum tree depth & 17 \\
    elite size & 5 \\
    initialisation &  position-independent grow \\
    maximum initial tree depth & 10 \vspace{0.3em}\\
    \hline
    \end{tabular*}
    \label{tab:sysParams}
\end{table}

Table~\ref{tab:sysParams} shows the system-specific parameters for PushGP and G3P, and the general parameters that are used in both systems. The ``runs per problem" parameter refers to the number of independent evolutionary runs that were conducted for each problem and experimental configuration. The PushGP system uses the uniform mutation by addition and deletion (UMAD) mutation operator \citep{Helmuth18UMAD}. This UMAD operator works with a 0.1 mutation rate. For G3P, we use subtree mutation and crossover, with a crossover probability of 0.95.
The initialization for G3P is position-independent grow \citep{Fagan16PI}. 
We use grammars based on those provided by the PonyGE2 framework with small adjustments to make them better comparable to the PushGP instructions. 
Across both systems and all runs, we use the same training and testing data. 
Whilst this might be questionable in terms of generalization to other data splits, we believe that seeing how the systems fare when provided with the same exact data is valuable. Additionally, this enables us to track which cases are used in a down-sample regularly across all runs over the course of a evolutionary process (see Sec.~\ref{sec:caseComp}).

\subsection{Evaluation and Generation Limits} \label{sec:genLimits}

In order to make a fair comparison between methods that perform different numbers of program executions per generation, we use the recommendation of the PSB2 benchmark suite to limit each GP run to 60 million program executions \citep{helmuth2021psb2}. Since program executions typically take up the majority of the computational requirements of a GP run, this ensures runs receive similar amounts of computation regardless of whether they use down-sampling. In standard runs using all training cases, the 60 million executions are used by at most 300 generations of a population size of 1000 individuals evaluated on 200 cases. With random down-sampling, we increase the maximum number of generations by the same factor as the down-sampling. For example, if one tenth of the training data is used in each sample, we can run evolution for ten times the number of generations while keeping the number of individual program executions constant.

More generally, if we let $G$ be the maximum number of generations for a run using all training cases, we allow our random down-sampling runs a limit of $\hat{G}$ generations where $\hat{G}$ is given by
$$\hat{G} = \dfrac{G}{r},$$
where $r$ is the down-sample rate.
For informed down-sampled lexicase selection the generational limit is calculated by
$$\hat{G} = \dfrac{G}{r + \frac{\rho(1-r)}{k}},$$
where $\rho$ is the parent sampling rate and $k$ is the parameter for the scheduled case distance computation.
The exact generational limits for each experimental configuration are shown in Table~\ref{tab:hyperparamSweep}.\footnote{As our implementations evaluate the fitness of individuals in the parent sample twice, we run the IDS with sparse information runs for slightly ($<40$) fewer generations to compensate the additional computational effort.} Although not considered in this work, there is a nominal cost to the Hamming distance operation between cases, however, this operation is dominated by the cost of performing the selection procedure, which is not considered in this work nor in prior work. In fact, lexicase selection has a time complexity of $\mathcal{O}(n^2 c)$ \citep{helmuth_population_2022}, where the Hamming distance operation has complexity $\mathcal{O}(c^2 n)$, where $c$ is the number of cases, and $n$ is the number of individuals. When $c$ is decreased due to down-sampling, the lexicase selection procedure becomes cheaper, partially accounting for the added cost of computing the hamming distance.

\subsection{Experimental Configurations}
We explore 11 different lexicase selection configurations for each problem: standard lexicase selection (Lex), 
random down-sampled lexicase selection (Rnd), IDS lexicase selection with full information, as well as three sparse information configurations.
To better match previous literature, all down-sampling methods were performed both with $r\in\{0.05; 0.1\}$. 

Table~\ref{tab:hyperparamSweep} shows the configurations of the different runs performed in this work. These runs, due to different generational computational costs, have different generational limits as explained in section~\ref{sec:genLimits}.

Full information down-sampling is simply using a parent sample rate of $1$, which means that the distances between the training cases are determined by the performance of all parents on each test case. With this, the quality of the distance metric between two cases is not limited by the parent sampling or generational gaps we are using to reduce computational load. Full information down-sampling is included as a control experiment to compare with using all cases for selection in standard lexicase selection. 
Note that all treatments using full information down-sampling ran for the same number of generations as standard lexicase selection because full information down-sampling requires \textit{all} parents to be evaluated on \textit{all} training cases.   

Finally, the six informed down-sampling methods we have chosen for this work include, for both the 0.05 and 0.1 down-sample rate ($r$), 0.01 parent sampling ($\rho$) rate with a few different distance calculation scheduling ($k$) parameters. Through a set of preliminary experiments, the value of $\rho = 0.01$ for the parent sampling rate was determined to be effective while not resulting in too many extra program executions\footnote{As we are trying to approach the computational savings of random down-sampled lexicase selection, the smaller the value of $\rho$, the better. We found that the relatively small value of $\rho = 0.01$ resulted in sampling that was good enough to determine the joint case information.}. In conjunction, these hyper-parameters mean that every $k$ generations, 10 parents are used to determine the distances between \emph{all} training cases, where $k\in\{1, 10, 100\}$. 

\renewcommand{\arraystretch}{1.35}
\begin{table}
    \caption{Different settings conducted in our experiments for standard lexicase selection (Lex), random down-sampled lexicase selection (Rnd) and informed down-sampled lexicase selection (IDS).
    The variable $r$ denotes the down-sampling rate, $\rho$ is the parent sampling rate, $k$ is generational interval at which we update the distance matrix and $\hat{G}$ specifies the maximum number of generations.}
    \tabcolsep=0.17cm
    \begin{tabular}{l|c|c|cccc|c|cccc}
    \hline
    \textbf{Method} & \textbf{Lex} & \textbf{Rnd}  & \multicolumn{4}{c|}{\textbf{IDS}} & \textbf{Rnd} &  \multicolumn{4}{c}{\textbf{IDS}}\\
    \hline
    $r$ & - & 0.05 & \multicolumn{4}{c|}{0.05} & 0.1 & \multicolumn{4}{c}{0.1} \\
    $\rho$ & - & - & 1 & 0.01 & 0.01 & 0.01 & - & 1 & 0.01 & 0.01 & 0.01 \\
    $k$ & - & - & 1 & 1 & 10 & 100 & - & 1 & 1 & 10 & 100\\
    $\hat{G}$ & 300 & 6000 & 300 & 5042 & 5888 & 5988 & 3000 & 300 & 2752 & 2973 & 2997\\
    \hline
\end{tabular}\label{tab:hyperparamSweep}
\end{table}

\section{Results and Discussion}\label{sec:Results}

\subsection{Informed Down-Sampling Improves Problem-solving Success} 

Tables~\ref{tab:push_results} and~\ref{tab:ge_results} show the success rates for PushGP and G3P respectively on the chosen program synthesis benchmark problems for different parameter configurations. 
We define success rate as the number of runs that result in a program that passes the complete training set, as well as the entire unseen test set. 

For random down-sampling and IDS, we measured solutions on only the down-samples during the actual run.  As such, we execute these runs to the maximum generational limit and then conduct a post hoc analysis to see if any solutions passed all of the training cases.
If so, this is the solution that we then evaluate on the unseen test set to determine whether it generalizes or not.

\renewcommand{\arraystretch}{1.35}
\definecolor{lightgray}{gray}{0.9}
\newcommand{\sig}[1]{\textbf{#1}}

\afterpage{
\begin{landscape}
\begin{table}[h!]
\centering
\vspace*{30mm}
\caption{Number of generalizing solutions (successes) out of 100 runs achieved by PushGP on the test set. For each benchmark problem, we highlight in \textbf{bold} the best success rate at each of the down-sample sizes. Problem names in \textbf{bold} are those where an informed down-sampling run outperformed random at \emph{both} down-sample rates on that problem. Problem names that are \underline{underlined} are those where a random down-sampling run outperformed an informed down-sampling run at both down-sample rates.
Asterisks signify results that are significantly better than random down-sampling \emph{at the same down-sample size}. Standard lexicase selection was not included in our statistical analyses, as IDS is presented to improve upon random down-sampling at a fixed down-sample size. We performed significance analysis with a two proportion z-test and Bonferroni-Holm correction. Shown with * are those significant at the $\alpha=0.1$ level, ** the $\alpha=0.05$ level, and *** the $\alpha=0.01$ level.
} 
\label{tab:push_results}
\renewcommand{\arraystretch}{1.3}
\begin{tabular}{l|c|ccccc|ccccc}
\hline
  \multicolumn{1}{l|}{\textbf{Method}} & \textbf{Lex} &  \textbf{Rnd}&  \multicolumn{4}{c|}{\textbf{IDS}}&  \textbf{Rnd} & \multicolumn{4}{c}{\textbf{IDS}} \\
  \multicolumn{1}{l|}{\textbf{ $r$ }} & \textbf{-} &   \multicolumn{5}{c|}{\textbf{0.05}} & \multicolumn{5}{c}{\textbf{0.1}} \\
  \multicolumn{1}{l|}{\textbf{$\rho$}} & \textbf{-} &  \textbf{-}& \textbf{1} & \textbf{0.01} & \textbf{0.01} & \textbf{0.01} & \textbf{-}& \textbf{1} & \textbf{0.01} & \textbf{0.01} & \textbf{0.01}  \\
  \multicolumn{1}{l|}{\textbf{$k$}} & \textbf{-} &  \textbf{-}& \textbf{1} & \textbf{1} & \textbf{10} & \textbf{100} & \textbf{-}& \textbf{1} & \textbf{1} & \textbf{10} & \textbf{100}  \\
\hline
\hline

 \sig{Count Odds} &     24 &    25 &    43*** &    99*** &    \sig{100***} &    98*** &    26 &   55*** &    95*** &    \sig{99***} &    97***  \\

\rowcolor{lightgray}
 \textbf{Find Pair} &    5 &    27 &    9 &    32 &    32 &    \textbf{36} &    15 &    7 &    19 &    19 &    \textbf{21}    \\

\sig{Fizz Buzz} &    13 &    64 &    2 &    85*** &   94*** &    \sig{95***} &    45 &    3 &    75 &    78* &    \sig{81**}    \\ 

\rowcolor{lightgray}
\sig{Fuel Cost} &     41 &    72 &    1 &    83 &   \textbf{85} &    83 &    \textbf{76} &    7 &    69 &    72 &    70    \\ 

 \sig{GCD}  &     20 &    74 &    4 &    \textbf{76} &    67 &    69 &    54 &    6 &    56 &    \textbf{63} &    62    \\

\rowcolor{lightgray}
 Grade &    0 &    0 &    0 &    0 & \sig{1} &    0 &    \sig{1} &    0 &  0 &    \sig{1} &    \sig{1}    \\

 \sig{Scrabble Score} &    8 &    8 &    6 &    69*** &    64*** &    \sig{75***} &    16 &    9 &    55*** &    \textbf{74***} &    64***    \\

\rowcolor{lightgray}

\underline{\smash{Small or Large}} &     34 &    \textbf{93} &    37 &    69 &    69 &    69 &    \textbf{69} &    39 &    60 &    66 &    54   \\ 
\hline
\end{tabular}
\end{table}
\end{landscape}
}

\afterpage{
\begin{landscape}
\begin{table}[h!]
\centering
\vspace*{30mm}
\caption{Number of generalizing solutions (successes) out of 100 runs achieved by G3P on the test set. For all configurations studied, we report success rates based on 100 runs. For each benchmark problem, we highlight in \textbf{bold} the best success rate at each of the down-sample sizes. Problem names in \textbf{bold} are those where an informed down-sampling run outperformed random at \emph{both} down-sample rates on that problem. Problem names that are \underline{underlined} are those where a random down-sampling run outperformed an informed down-sampling run at both down-sample rates.
Asterisks signify results that are significantly better than random down-sampling \emph{at the same down-sample size}. Standard lexicase selection was not included in our statistical analyses, as IDS is presented to improve upon random down-sampling at a fixed down-sample size. We performed significance analysis with a two proportion z-test and Bonferroni-Holm correction. Shown with * are those significant at the $\alpha=0.1$ level, ** the $\alpha=0.05$ level, and *** the $\alpha=0.01$ level.
}\label{tab:ge_results}
\renewcommand{\arraystretch}{1.3}
\begin{tabular}{l|c|ccccc|ccccc}
\hline
  \multicolumn{1}{l|}{\textbf{Method}} & \textbf{Lex} &  \textbf{Rnd}&  \multicolumn{4}{c|}{\textbf{IDS}}&  \textbf{Rnd} & \multicolumn{4}{c}{\textbf{IDS}} \\
  \multicolumn{1}{l|}{\textbf{$r$ }} & \textbf{-} &   \multicolumn{5}{c|}{\textbf{0.05}} & \multicolumn{5}{c}{\textbf{0.1}} \\
  \multicolumn{1}{l|}{\textbf{$\rho$}} & \textbf{-} &  \textbf{-}& \textbf{1} & \textbf{0.01} & \textbf{0.01} & \textbf{0.01} & \textbf{-}& \textbf{1} & \textbf{0.01} & \textbf{0.01} & \textbf{0.01}  \\
  \multicolumn{1}{l|}{\textbf{$k$}} & \textbf{-} &  \textbf{-}& \textbf{1} & \textbf{1} & \textbf{10} & \textbf{100} & \textbf{-}& \textbf{1} & \textbf{1} & \textbf{10} & \textbf{100}  \\
\hline
\hline

Count Odds &    65 &    \textbf{66} &    45 &    53 &    62 &    63 &    67 &    58 &    60 &    58 &    \textbf{72}    \\ 

\rowcolor{lightgray}
 Find Pair &    0 & 0 & 0 & \textbf{1} & 0 & 0 & \textbf{1} & 0 & 0 & \textbf{1} & 0     \\ 

 \textbf{Fizz Buzz}&   62 &    83 &    50 &    84 &    78 &    \textbf{85} &    78 &    53 &    81 &    \textbf{89} &    72    \\ 

 \rowcolor{lightgray}
 Fuel Cost&    33 &    \textbf{34} &    17 &    28 &    27 &    29 &    29 &    21 &    21 &    25 &    \textbf{33}    \\ 
 
 GCD  & 0 & \textbf{1} & 0 & 0 & 0 & \textbf{1} & 0 & 0 & 0 & 0 & 0    \\

\rowcolor{lightgray}

 \textbf{Grade} &    36 &    39 &    29 &    51 &    \textbf{57*} &    44 &    44 &    37 &    46 &    \textbf{51} &    48    \\

 Scrabble Score &    6 & 10 & 1 & \textbf{11} & 10 & 10 & \textbf{14} & 0 & 6& 3 & 3  \\

\rowcolor{lightgray}
 \textbf{Small or Large}  &    41 &    52 &    49 &    54 &    \textbf{63} &    \textbf{63} &    59 &    52 &    57 &    55 &    \textbf{63}    \\

\hline
\end{tabular}
\end{table}
\end{landscape}
}


Overall, at least one configuration of informed down-sampling resulted in the greatest success rate for 6/8 problems solved by PushGP, with 3/8 of them being statistically significant. 
With PushGP, random down-sampling significantly outperformed informed down-sampling on one problem. 
For G3P, at least one configuration of informed down-sampling produced the greatest success rate for 3/8 problems, with 1/8 being significant.
Random down-sampling produced the greatest success rate on two problems with G3P; though, neither instance was statistically significant.
While we did not make direct comparisons with standard lexicase, we observed greater success rates with both random and informed down-sampling than with standard lexicase.

Our data show that informed down-sampling by farthest first traversal can improve problem-solving success beyond random down-sampling lexicase, which is the current state-of-the-art for program synthesis \citep{helmuth_benchmarking_2020}.
As is often the case, however, the benefits of informed down-sampling varied by problem and GP representation. 
With PushGP informed down-sampling dramatically improved problem-solving success rates for the Count Odds, Fizz Buzz, and Scrabble Score problems, but decreased success on Small Or Large.
G3P less clearly benefited from informed down-sampling; though, we did observe minor, configuration-dependent improvements in overall problem-solving success.  

This is clear as informed down-sampling at all configurations ensures that close to if-not-all 100 runs successfully generalize to the held out test set. This and similar results hint that while randomly down-sampled lexicase selection works well usually, there are some problems where important cases might be being dropped out, resulting in a similar performance to standard lexicase selection 
despite the increased search generations. Informed down-sampling has the ability to improve success rates both when random down-sampling improves upon standard lexicase selection, and when it does not.

Only one configuration of G3P resulted in a significant improvement on random down-sampling at the same down-sample rate. For the Grade problem at the $0.05$ down-sample rate, we see significantly more successes when using IDS with $\rho = 0.01$ and $k = 10$. For this problem, using this informed down-sample configuration resulted in 57\% of the runs yielding a generalizing solution, where, using random down-sampling resulted in only 39\% of the runs yielding a success.
The fact that only a single configuration of IDS resulted in a significant improvement suggests that the problem-solving benefits of using IDS are representation- and problem-dependent, motivating future work to continue improving IDS to achieve more universal improvements to problem-solving success. 

We have a number of hypotheses explaining this improved performance. The first of these is that the informed down-sampling procedure increases the number of specialists (individuals exceptional on a few cases, but have a high \emph{total} error) that survive over the course of evolutionary time. These individuals could be better maintained with IDS as the cases they are exceptional on are still placed in the down-samples throughout evolution, preventing them from being lost as could happen when randomly down-sampling.

Another hypothesis for IDS's improved performance is that it reduces the computation used to evaluate individuals on synonymous cases.
When two cases are fully synonymous, all individuals that solve one case solve the other as well. When using lexicase selection, having both of these cases in the down-sample would result in little difference in the probability of selecting each individual compared to having only one case in the down-sample. After one of the two cases has been used to filter the pool of candidate solutions, the other will have no filtering pressure because all remaining individuals perform identically on the synonymous cases. Having a synonymous case \textit{does} increase the chance that one of the two cases appears earlier in the shuffled case ordering, producing a minor (though perhaps undesired) change in selection probability. Synonymous (or near synonymous) cases additionally take spots in the down-sample that cannot be allocated to other, more-informative cases. When using IDS, we ensure that the first few cases added to the down-sample measure relatively different behaviors. This may allow IDS to select a larger variety of individuals than random down-sampling, instead approximating the variety that could be selected by full lexicase selection.


Random down-sampling outperformed informed down-sampling (across both down-sampling levels) on only one problem (\emph{Small or Large}) for PushGP, and none for G3P. For \emph{Small or Large} with PushGP, we see that the worse performance with informed down-sampling can be attributed to a lower generalization rate (and not worse performance on the training sets). The generalization rates can be found in Appendix Table~\ref{tab:push_generalization} for PushGP and Appendix Table~\ref{tab:ge_generalization} for G3P. Future work should explore the effect that informed down-sampling has on generalization in more depth.

Appendix~\ref{appendix:sizes} shows the distributions of program size across the runs that were performed for this work. Despite the larger generational limit, there does not appear to be any significant bloating that occurs during our down-sampled runs when compared to our lexicase selection runs, despite the 10x-20x larger generational limit. This provides evidence that the method we used to ensure that all programs use similar computational resources is valid, as there are not drastic differences in program size (and thus evaluation cost). Furthermore, we employ strict evaluation limits for our PushGP runs, where programs are cut off after they reach the step limit of 200 steps. For G3P, we have a strict looping limit of 1000 iterations.

Although we introduce a variety of hyperparameters in this work, we found the most consistent improvement yielded by choosing $r=0.05, \rho=0.01, k=10$. As such, we recommend these hyperparameters as a starting point for future investigation.

\subsection{Using Smaller Informed Down-Samples Tends to Improve Success Rates}

In general, our IDS runs at a 0.05 down-sample rate have a higher success rate than their equivalent counterparts at the 0.1 down-sample rate. This difference is likely due to the fact that the runs at a 0.1 down-sample rate have a substantially lower generational limit, meaning that we are exploring a smaller portion of the space of possible solution programs. With 200 training cases, our down-sample contains 10 and 20 cases respectively for the 0.05 and 0.1 down-sample rates. A possible reason for the improved performance at 0.05 is that a larger proportion of these cases are indeed our distinct, or informative, cases. Note that once the Farthest First Traversal process selects a representative case for every synonymous group in the down-sample, every remaining solution's minimum distances to the current sample will be equal to 0, so the selections are performed randomly to fill the rest of the cases. Since we are using the same problems, with the same number of behavioral niches, the runs with 20 cases in the down-sample will have more synonymous cases in the down-sample. 
As such, the larger sample sizes do not necessarily result in substantially more informative samples of training cases than smaller sample sizes. 
We will analyze the specific cases that compose the down-samples in section~\ref{sec:caseComp}.

The exceptions to this trend are the full information down-sampling runs. For these runs, the larger down-samples tend to perform better. This result is likely due to the fact that the generational limit was set to 300 for both sampling levels (as they both evaluate all individuals on all test cases), and so having a smaller down-sample size would not change the number of evaluations. With more cases in the sample, the GP method can take into account more information when performing selection, which could result in more informed search. The magnitude of the differences for success rate across sample size for the full IDS runs suggests that there are diminishing returns for including more cases in the sample.

\subsection{Informed Down-Sampling Automatically Discovers Important Training Cases}\label{sec:caseComp}

To gain a deeper insight into how IDS composes down-samples, we visualized the set of training cases selected for informed down-samples over an evolutionary run. 

Figures~\ref{fig:push_downsample_comp_1} and~\ref{fig:ggg_downsample_comp_1} show the composition of down-samples for every problem at every generation using PushGP (Fig.~\ref{fig:push_downsample_comp_1}) and G3P (Fig.~\ref{fig:ggg_downsample_comp_1}) with down-sample rate $r=0.05$. We present results for a full information configuration ($\rho=1$ and $k=1$) as well as a sparse information configuration ($\rho=0.01$ and $k=10$). We chose to analyze both a full information and sparse information run in order to see whether our sparse information configurations are finding the same training cases to be informative as if we had used all parents to evaluate the distances between training cases.

The plots show how often certain training cases are included in the down-sample at every generation, averaged over all active runs. Each row represents a case in the training data, ordered by its position in the training set. The training sets used were generated by first adding some human-expert defined edge cases, and filling the rest with cases that were randomly generated by an function that already implements our desired program (oracle function). For each figure, there is a single marker on the $y$-axis that shows where exactly the expert-case cutoff for the training set was. Thus, the rows above the marker in the visuals are representing cases that humans determined to be important based on the problem definition. 

Brighter colors imply that a case is included more often, darker colors imply a lower number of inclusions.

\begin{figure}[!ht]
    \centering
    \includegraphics[width=0.85\textwidth]{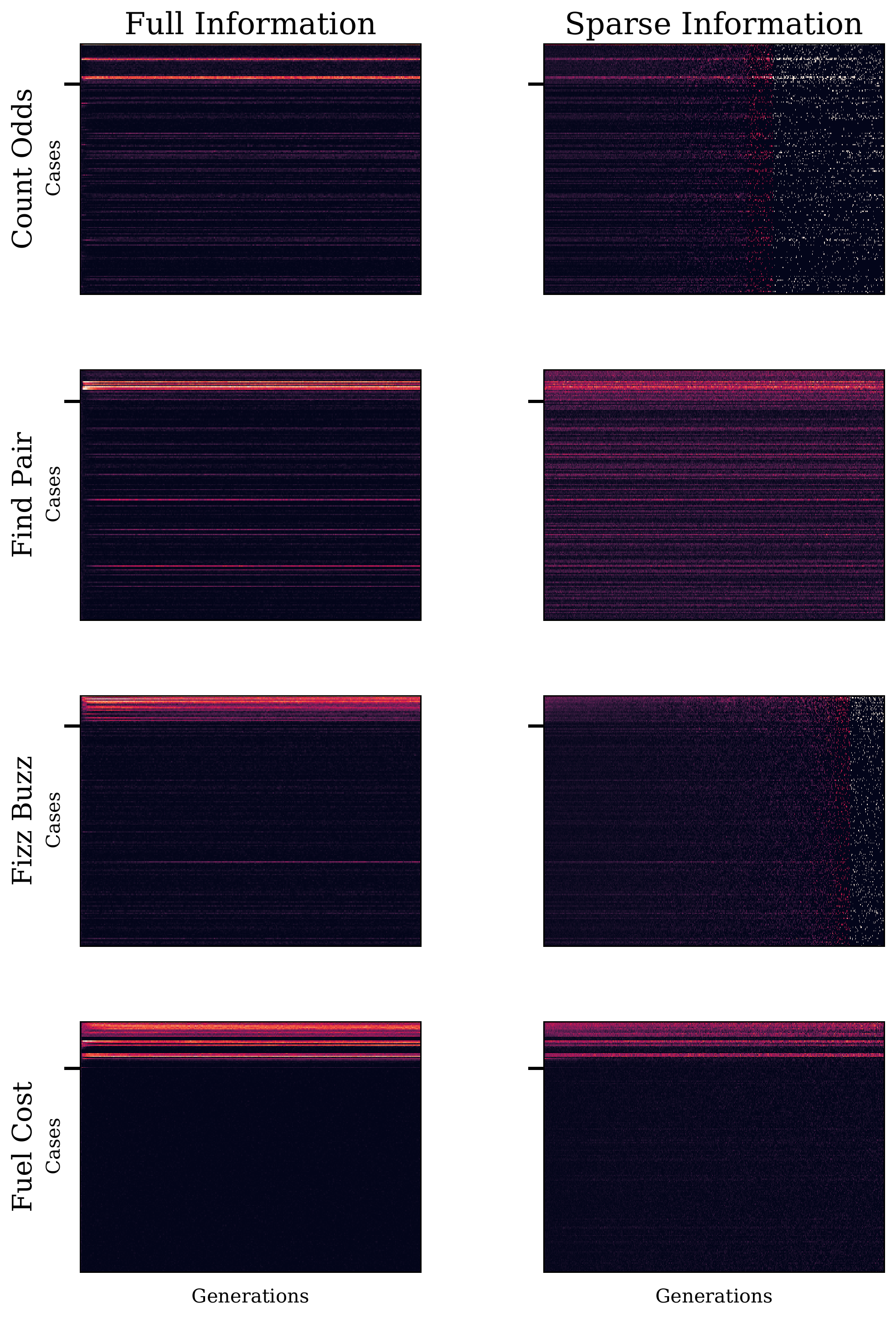}
    \caption{Down-sample composition over generations for PushGP with 0.05 down-sample rate for a full information ($\rho=1$ and $k=1$) and a sparse information configuration ($\rho=0.01$ and $k=10$). }
    \label{fig:push_downsample_comp_1}
\end{figure}

\begin{figure}[!ht]
\ContinuedFloat
    \centering
    \includegraphics[width=0.85\textwidth]{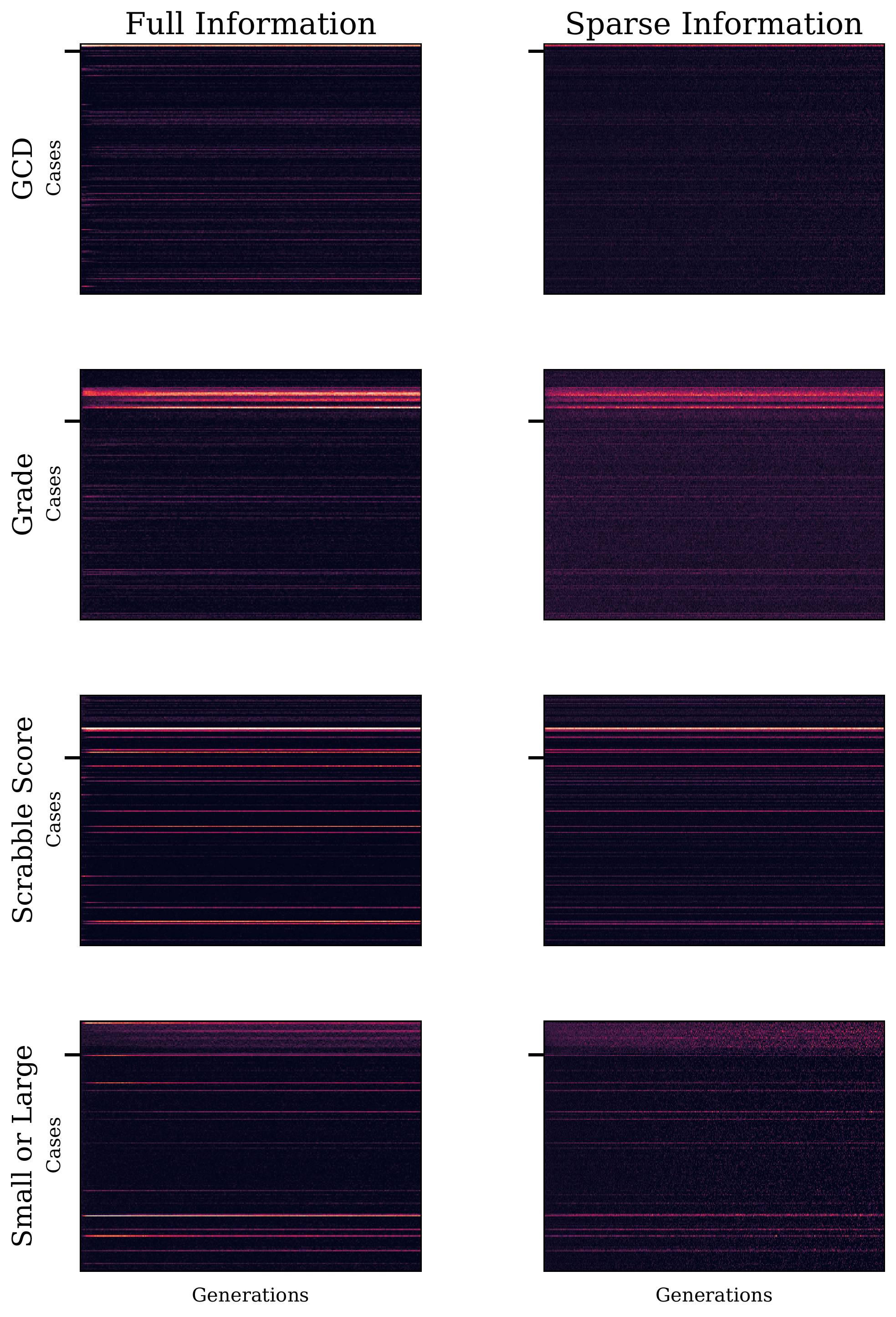}
    \caption{Continued.}
    \label{fig:push_downsample_comp_2}
\end{figure}

\begin{figure}[!ht]
    \centering
    \includegraphics[width=0.85\textwidth]{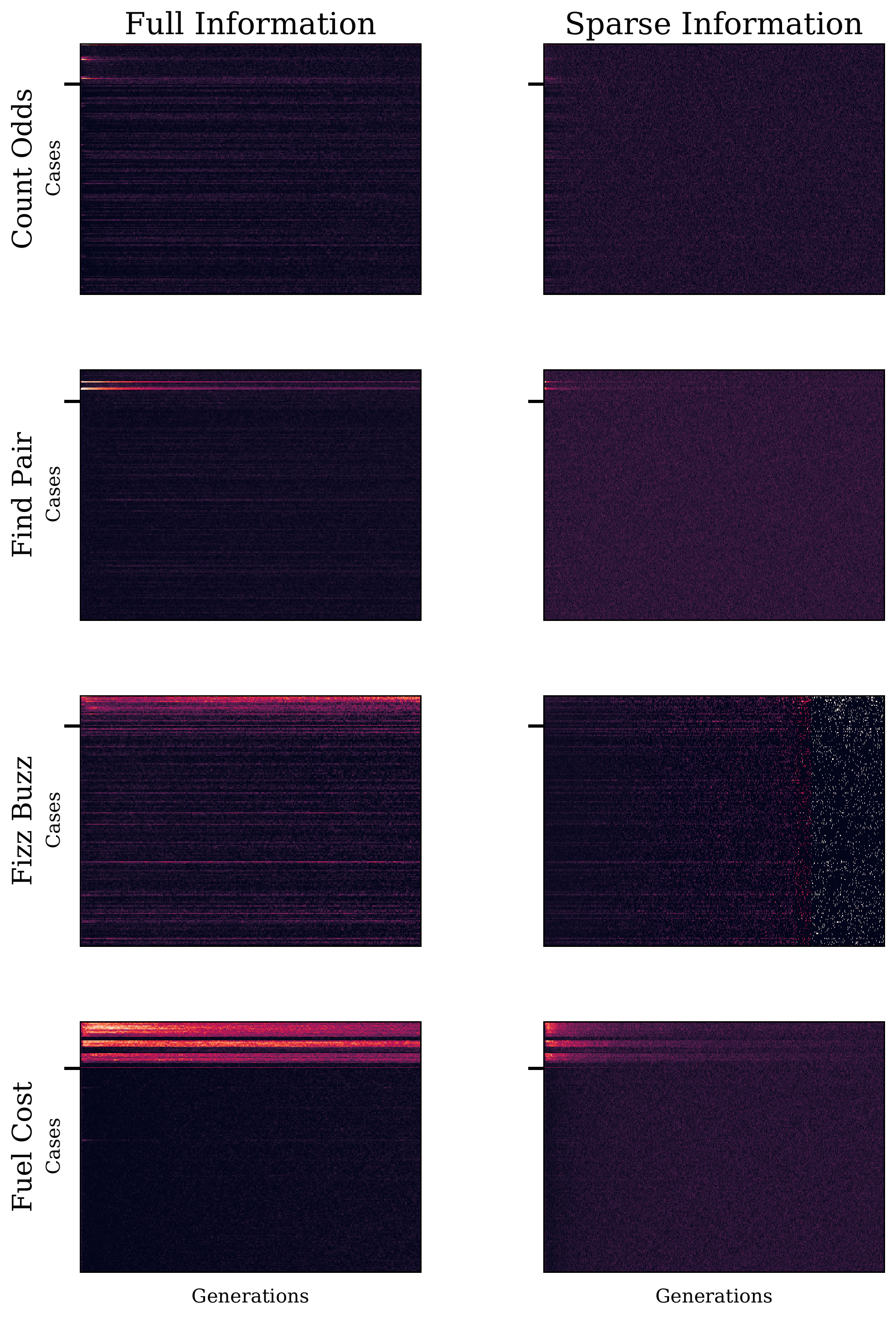}
    \caption{Down-sample composition over generations for G3P with 0.05 down-sample rate for a full information ($\rho=1$ and $k=1$) and a sparse information configuration ($\rho=0.01$ and $k=10$). \\ }
    \label{fig:ggg_downsample_comp_1}
\end{figure}

\begin{figure}[!ht]
\ContinuedFloat
    \centering
    \includegraphics[width=0.85\textwidth]{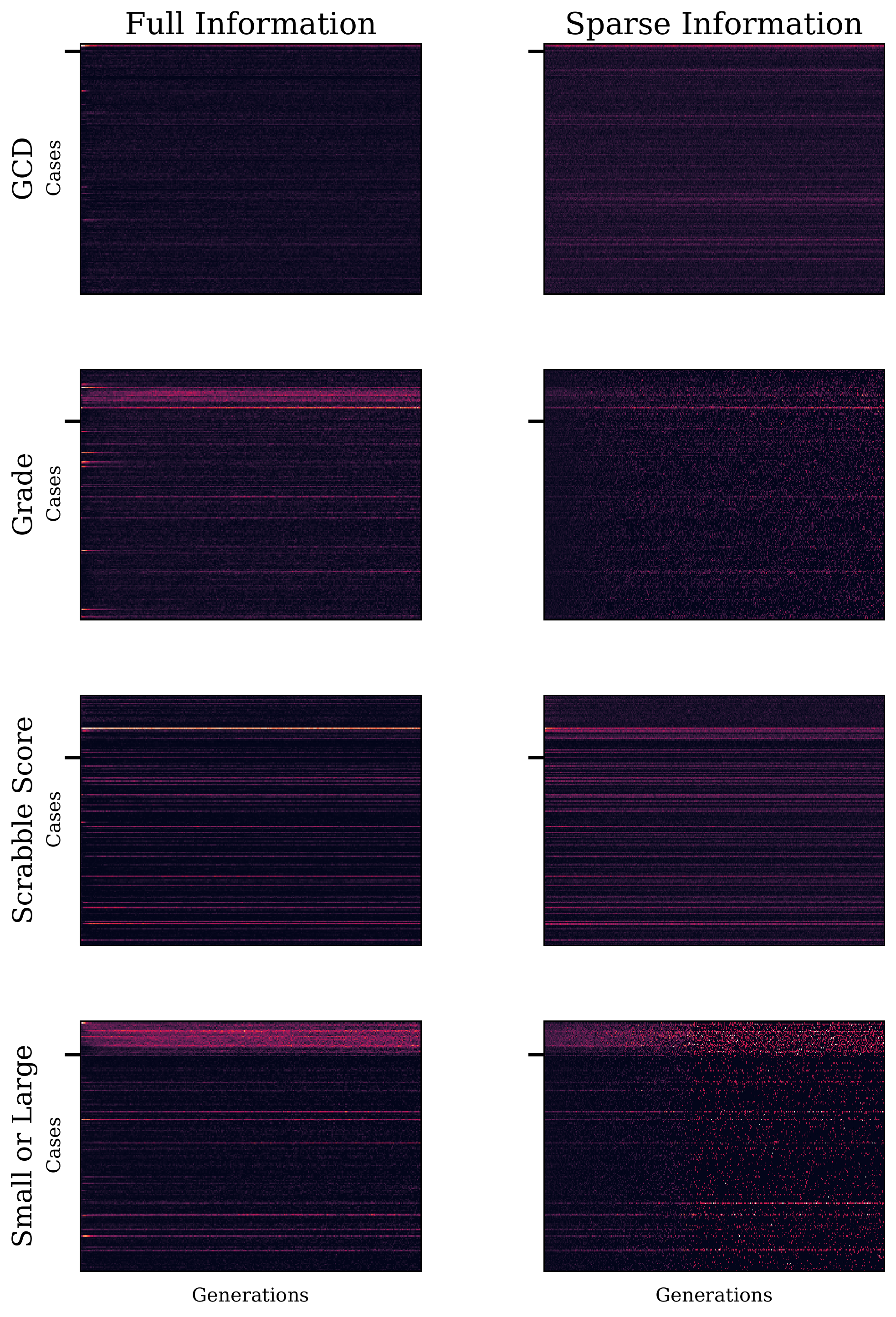}
    \caption{Continued.}
    \label{fig:ggg_downsample_comp_2}
\end{figure}

For PushGP (Figure~\ref{fig:push_downsample_comp_1}), we see that the configurations with sparse information often include the same cases in the down-sample as the runs with full information. This result means that by using a parent sampling rate of $\rho = 0.01$ and a case distance evaluation schedule parameter of $k=10$, we can significantly reduce the number of evaluations needed to calculate distances between cases, while still maintaining a good approximation to the ground truth (full information, where we use all parents every generation to calculate distances). However, the composition for our sparse information runs are slightly more noisy than that for full information, suggesting that using parent sampling could introduce some extra stochasticity to the down-sample creation process.

For all studied benchmark problems, we see that IDS has a strong bias toward specific training cases that are included substantially more often in the down-sample. 
These selected training cases are mainly consistent with the human-defined edge cases that exist at the beginning of the training set. This result shows that informed down-sampling is indeed often finding the same cases to be informative as those that a human expert would, without any knowledge of the problem definition.
However, with IDS, we can draw further comparisons of informativeness within this expert-defined groups of cases. This can be seen as some cases are selected more often that others \emph{within} the first several cases.

We then look at the labels of the specific training cases that are found to be important. We see that these training cases make sense to be included more often than others in the down-samples. Note that the labels of the specific training cases are not included in the plots for simplicity, but can be queried based on their specific index in the data sets provided in our code implementation.

For example, for the \emph{Small or Large} problem, cases around the decision boundaries as well as numbers between $0$ and $1000$ are more often included. 
For the \emph{Grade} problem, those edge cases with very close decision boundaries are included while the ones with far away boundaries are not taken into account for the down-sample.
For \emph{Fuel Cost}, nearly all of the human defined edge cases are found to be important, while for the \emph{GCD} problem the first two cases in particular make it in nearly every down-sample, while the rest are selected less often.

For the \emph{Scrabble Score} problem, we see that the first edge cases, which specify the score for each letter, does not seem to be informative at all. This result is not surprising, as this information is already available to PushGP through a vector with these scores on the vector stack. However, the three edge cases after them with empty strings and special characters as input are frequently included.
For \emph{Count Odds}, the edge cases denoting empty lists, or lists with zero or a single odd number were found to be important, indicating that those contain all the important information to learn what are odd and even numbers as well as how to handle a list.
For \emph{Fizz Buzz}, all edge cases seem important while for the \emph{Find Pair} problem only those edge cases with lists of length $3$ are consistently included. Those lists of length $2$ in the edge cases are represented in the down-sample less often.

Lastly, we see that the composition of the down-sample stays rather stable during the evolutionary run for the PushGP system, explaining why there is only a small difference in our experiments between calculating the distances every $k=1$ and $k=100$ generations (see Table~\ref{tab:push_results}).

For G3P (Fig~\ref{fig:ggg_downsample_comp_1}), we see similar results as with PushGP. However, for the problems that require iterative structures to be solved (\emph{Count Odds}, \emph{Find Pair}) we see that the down-sample quickly dissolves into random noise instead of any form of structure. This dynamic occurs despite the fact that the same edge cases as with PushGP are initially identified in the first few generations. This result is not surprising as finding iterative structures is known to be challenging for grammar-guided approaches, as such structures are difficult to be built step-by-step guided by the performance on a set of training cases \citep{sobania2020challenges, sobania2022comprehensive}. Another difference between the case compositions are that, while IDS with G3P tends to discover the same cases as those found with PushGP, their use is less consistent, resulting in lines that are more faint than those for PushGP. Both of these hypotheses could help explain the relatively worse improvement that IDS yields for G3P than for PushGP.

However, for the problems that require conditionals, like \emph{Small or Large} and \emph{Grade}, we see that the important cases are identified and used during evolution. This result is also reflected in the success rates compared to random down-sampling (see Table~\ref{tab:ge_results}).

Interestingly, IDS identifies many of the same cases as important for G3P as well as PushGP. This result suggests that the structure of the problem itself determines which cases are important rather than the considered representation. This dynamic makes IDS potentially useful across many different systems and approaches. 

Across both systems, we see that the important cases are determined and selected very early in evolution, despite the relatively uninformative environment at the beginning of evolution. At the beginning of the run, most cases are failed by all individuals, making it hard to distinguish between informative and uninformative cases. An option that could help address this issue would be to use raw fitness values to measure the distance between test cases. However, using this system would require the definition of a similarity function based on fitness values, which could introduce hyper-parameters that detract from the simplicity of our approach.

\section{Conclusion and Future work}\label{sec:conclusion}

In this work, we demonstrated a novel approach to constructing down-samples using runtime population statistics.
We found that changing the composition of down-samples to include cases that are more ``informative" helps improve problem-solving success given a fixed computational budget.
By replicating our experiments across multiple GP systems (PushGP and G3P), we found that the problem-solving benefits of informed down-sampling vary by search space and by program representation, warranting further study. 

We hypothesize that a down-sample's ``informativeness'' is linked to how distinct its constituent cases are from one another. 
Cases solved by the same subset of the population are likely testing for the same behavior, and thus need not be included in the down-sample at the same time. 
We verified that our approach to informed down-sampling selects distinct sets of important training cases (as defined by human experts).  
Cases that test for different behaviors likely maintain different behavioral groups of individuals, which could promote and maintain higher levels of diversity in the population.
Indeed, static analyses of informed down-sampling have shown that it is better capable of selecting specialists than random down-sampling~\citep{boldi_static_2023}.


This work is a first exploration into more systematic approaches to building down-samples for lexicase selection runs. As such, it opens many potential directions for future research. Due to the modular nature of the informed down-sampling system, different methods could be used for either the pairwise information measurement, or for the down-sample creation portions of the algorithm. An exploration into different down-sampling levels, and the effects levels have on the informational content of down-samples is also a promising direction for future work. Additionally, IDS introduces new hyperparameters for the parent sampling rate and generational schedule.
Future work will explore methods for reducing the number of new hyperparameters by automatically configurating them at runtime based on the problem and state of the GP search.  
Finally, even though there are reasons to believe that IDS and down-sampling in general work well with lexicase selection, there is nothing that ties them to a particular selection method.
We encourage future studies exploring the effects of IDS on other parent selection methods such as tournament selection (e.g., \citealt{boldi_problem_2023}). 
Finally, comparing the extent to which different down-sampling strategies blunt lexicase's ability to maintain specialists could also yield important insights into why informed down-sampling improves success rates as much as it does. 

\section{Acknowledgements}
This material is based upon work supported by the National Science Foundation under Grant No. 1617087. Any opinions, findings, and conclusions or recommendations expressed in this publication are those of the authors and do not necessarily reflect the views of the National Science Foundation.

This work was performed in part using high performance computing equipment obtained under a grant from the Collaborative R\&D Fund managed by the Massachusetts Technology Collaborative.

Parts of this research were conducted using the supercomputer Mogon and/or advisory services offered by Johannes Gutenberg University Mainz (hpc.uni-mainz.de), which is a member of the AHRP (Alliance for High Performance Computing in Rhineland Palatinate,  www.ahrp.info) and the Gauss Alliance e.V.

The authors would like to thank Anil Saini, Austin Ferguson, Cooper Sigrist, Constantin Weiser, Edward Pantridge, Jose Hernandez, Li Ding, the Members of the PUSH lab at Amherst College as well as the anonymous reviewers for discussions and suggestions that helped shape this work. 
\small

\bibliographystyle{apalike}
\bibliography{ecjsample} 

\newpage

\appendix

\section{Generalization Rates}


\begin{table}[h!]
\centering
\caption{Generalization rate for PushGP. These data indicate the proportion of the runs that passed the training set that also passed the held out test set.
\vspace{0.11cm}}\label{tab:push_generalization}
\renewcommand{\arraystretch}{1.3}
\begin{tabular}{l|c|ccccc|ccccc}
\hline
  \multicolumn{1}{l|}{\textbf{Method}} & \textbf{Lex} &  \textbf{Rnd}&  \multicolumn{4}{c|}{\textbf{IDS}}&  \textbf{Rnd} & \multicolumn{4}{c}{\textbf{IDS}} \\
  \multicolumn{1}{l|}{\textbf{ $r$ }} & \textbf{-} &   \multicolumn{5}{c|}{\textbf{0.05}} & \multicolumn{5}{c}{\textbf{0.1}} \\
  \multicolumn{1}{l|}{\textbf{$\rho$}} & \textbf{-} &  \textbf{-}& \textbf{1} & \textbf{0.01} & \textbf{0.01} & \textbf{0.01} & \textbf{-}& \textbf{1} & \textbf{0.01} & \textbf{0.01} & \textbf{0.01}  \\
  \multicolumn{1}{l|}{\textbf{$k$}} & \textbf{-} &  \textbf{-}& \textbf{1} & \textbf{1} & \textbf{10} & \textbf{100} & \textbf{-}& \textbf{1} & \textbf{1} & \textbf{10} & \textbf{100}  \\
\hline
\hline
 Count Odds& 1.00  &    0.96 &    0.98 &    0.99 & 1.00  &  0.99  &    0.96 &    1.00  &    0.98 &    0.99 &    0.99    \\ 
 \rowcolor{lightgray}

 Find Pair&   1.00  & 0.82 & 0.82 & 0.73 & 0.74 & 0.80 & 0.50 & 0.88 & 0.79 & 0.68 & 0.75      \\ 

  Fizz Buzz&   0.93 &    0.96 &   1.00  &    0.93 &    0.95 &    0.99 &  1.00  &  1.00  &    0.96 &    0.96 &    0.96    \\ 
 \rowcolor{lightgray}
  Fuel Cost&    1.00 &   1.00 &    1.00 &    0.99 &    0.99 &    0.99 &    1.00 &    1.00  &    1.00  &    1.00 &    1.00    \\ 
 GCD  & 0.91 & 0.93 & 1.00  & 0.93 & 0.83 & 0.87 & 0.82 & 0.75 & 0.80 & 0.89 & 0.87    \\
 \rowcolor{lightgray}
  Grade & - &   - &   - &   - & 1.00 & - &    1.00 &    - &    - &    1.00  &    1.00     \\ 

 Scrabble Score & 1.00 & 1.00 & 1.00 & 1.00 & 1.00 & 1.00 & 1.00 & 1.00 & 0.98 & 1.00 & 1.00  \\ 
\rowcolor{lightgray}
  
 Small or Large  &    0.71 &    0.95 &    0.80 &    0.78 &    0.74 &    0.71 &    0.81 &    0.77 &    0.69 &    0.73 &    0.64    \\

\hline
\end{tabular}
\end{table}

\begin{table}[h!]
\centering
\caption{Generalization rate for G3P. These data indicate the proportion of the runs that passed the training set that also passed the held out test set.
\vspace{0.11cm}}\label{tab:ge_generalization}
\renewcommand{\arraystretch}{1.3}
\begin{tabular}{l|c|ccccc|ccccc}
\hline
  \multicolumn{1}{l|}{\textbf{Method}} & \textbf{Lex} &  \textbf{Rnd}&  \multicolumn{4}{c|}{\textbf{IDS}}&  \textbf{Rnd} & \multicolumn{4}{c}{\textbf{IDS}} \\
  \multicolumn{1}{l|}{\textbf{ $r$ }} & \textbf{-} &   \multicolumn{5}{c|}{\textbf{0.05}} & \multicolumn{5}{c}{\textbf{0.1}} \\
  \multicolumn{1}{l|}{\textbf{$\rho$}} & \textbf{-} &  \textbf{-}& \textbf{1} & \textbf{0.01} & \textbf{0.01} & \textbf{0.01} & \textbf{-}& \textbf{1} & \textbf{0.01} & \textbf{0.01} & \textbf{0.01}  \\
  \multicolumn{1}{l|}{\textbf{$k$}} & \textbf{-} &  \textbf{-}& \textbf{1} & \textbf{1} & \textbf{10} & \textbf{100} & \textbf{-}& \textbf{1} & \textbf{1} & \textbf{10} & \textbf{100}  \\
\hline
\hline

 Count Odds&    0.94 &    0.96 &    0.96 &    0.88 &    1.00 &    0.96 &    1.00 &    0.92 &    0.95 &    0.91 &    0.95   \\ 
\rowcolor{lightgray}
 Find Pair&    - & - & - & 1.00 & - & - & 1.00 & - & - & 1.00 & -      \\ 

 Fizz Buzz&   0.79 &    0.87 &    0.85 &    0.84 &    0.78 &    0.85 &    0.83 &    0.82 &    0.82 &    0.89 &    0.73    \\ 

 \rowcolor{lightgray}
 Fuel Cost&    1.00 &    0.97 &    1.00 &    0.97 &    0.96 &    1.00 &    1.00 &    0.96 &    0.96 &    1.00 &    1.00    \\ 

  GCD  & - & 0.17 & - & - &- & 0.25 & - & - & - & - & -    \\ 

\rowcolor{lightgray}
 Grade &    0.42 &    0.45 &    0.50 &    0.53 &    0.59 &    0.45 &    0.47 &    0.54 &    0.47 &    0.54 &    0.49    \\ 

 Scrabble Score &    1.00 & 1.00 & 1.00 & 1.00 & 0.92 & 0.83 & 1.00 & - & 0.86 & 1.00 & 0.60  \\ 

 \rowcolor{lightgray}
 Small or Large  &    0.47 &    0.57 &    0.65 &    0.56 &    0.64 &    0.66 &    0.68 &    0.59 &    0.60 &    0.579 &    0.65   \\ 

\hline

\end{tabular}
\end{table}

\section{Average Program Size}\label{appendix:sizes}

\begin{figure}
  \centering
  \begin{subfigure}[b]{0.45\linewidth}
    \centering
    \includegraphics[width=\linewidth]{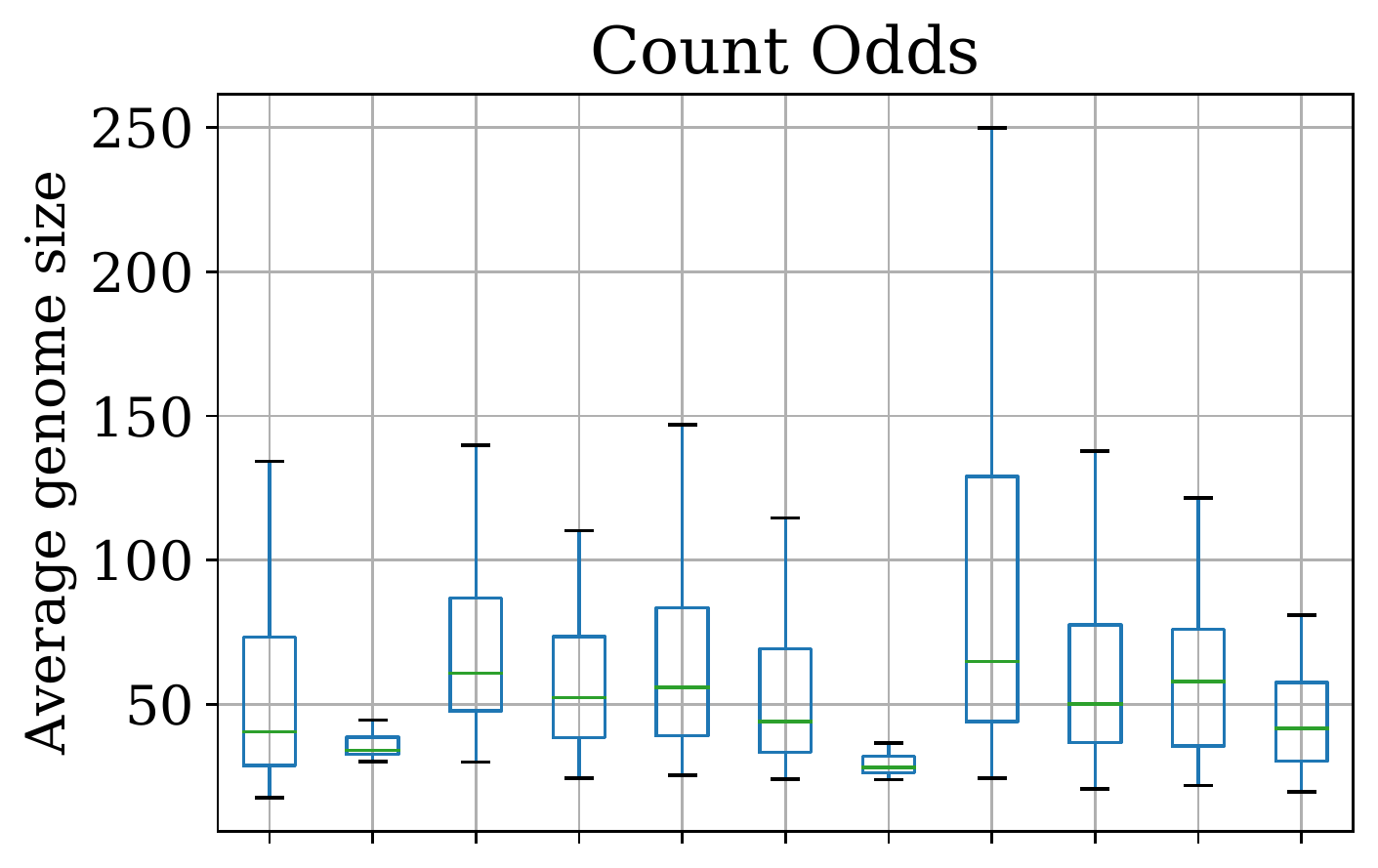}
    \label{fig:sub_push_size_1}
  \end{subfigure}
  \hfill
  \begin{subfigure}[b]{0.45\linewidth}
    \centering
    \includegraphics[width=\linewidth]{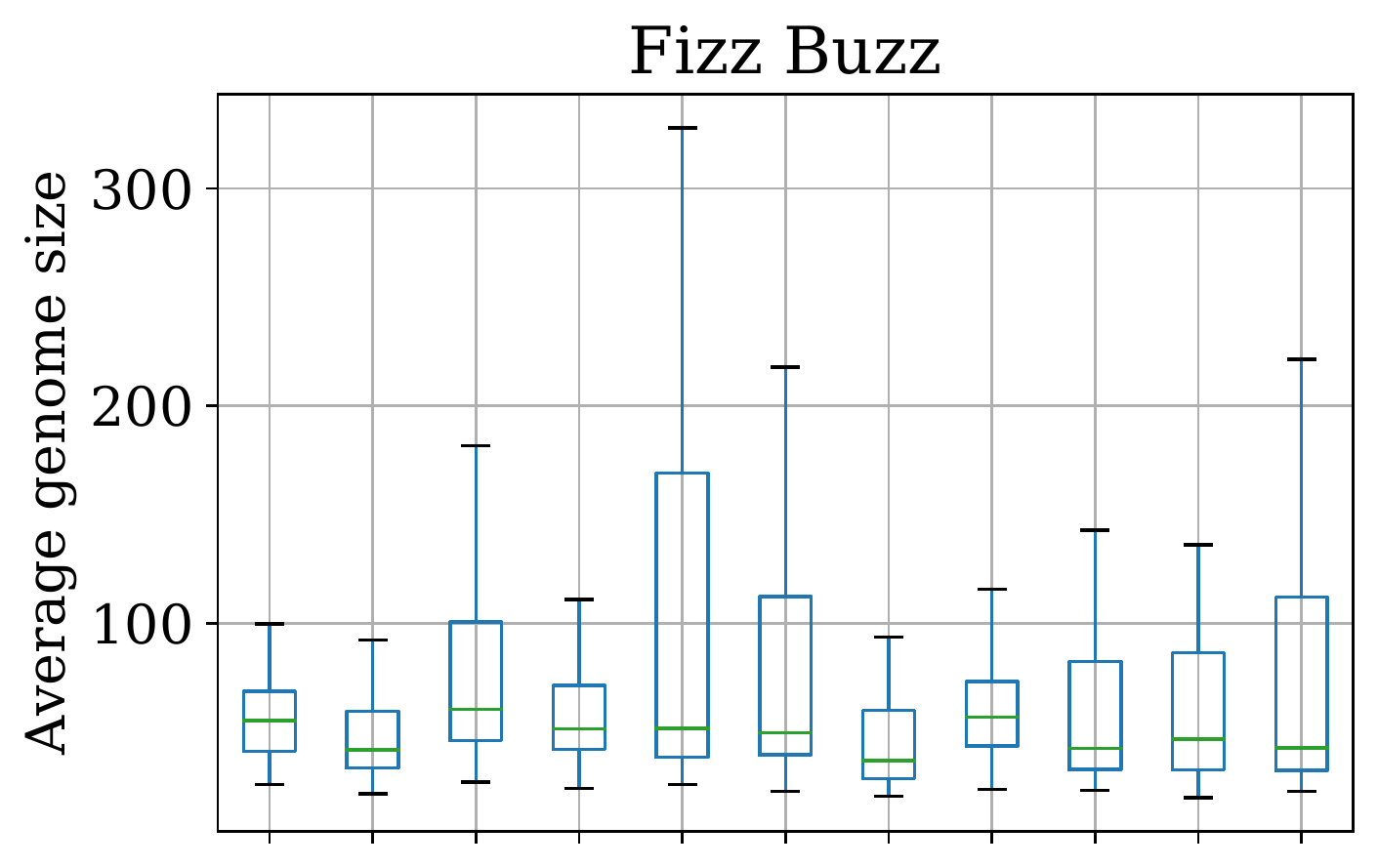}
    \label{fig:sub_push_size_2}
  \end{subfigure}
  
  \vspace{0.2cm}
  
  \begin{subfigure}[b]{0.45\linewidth}
    \centering
    \includegraphics[width=\linewidth]{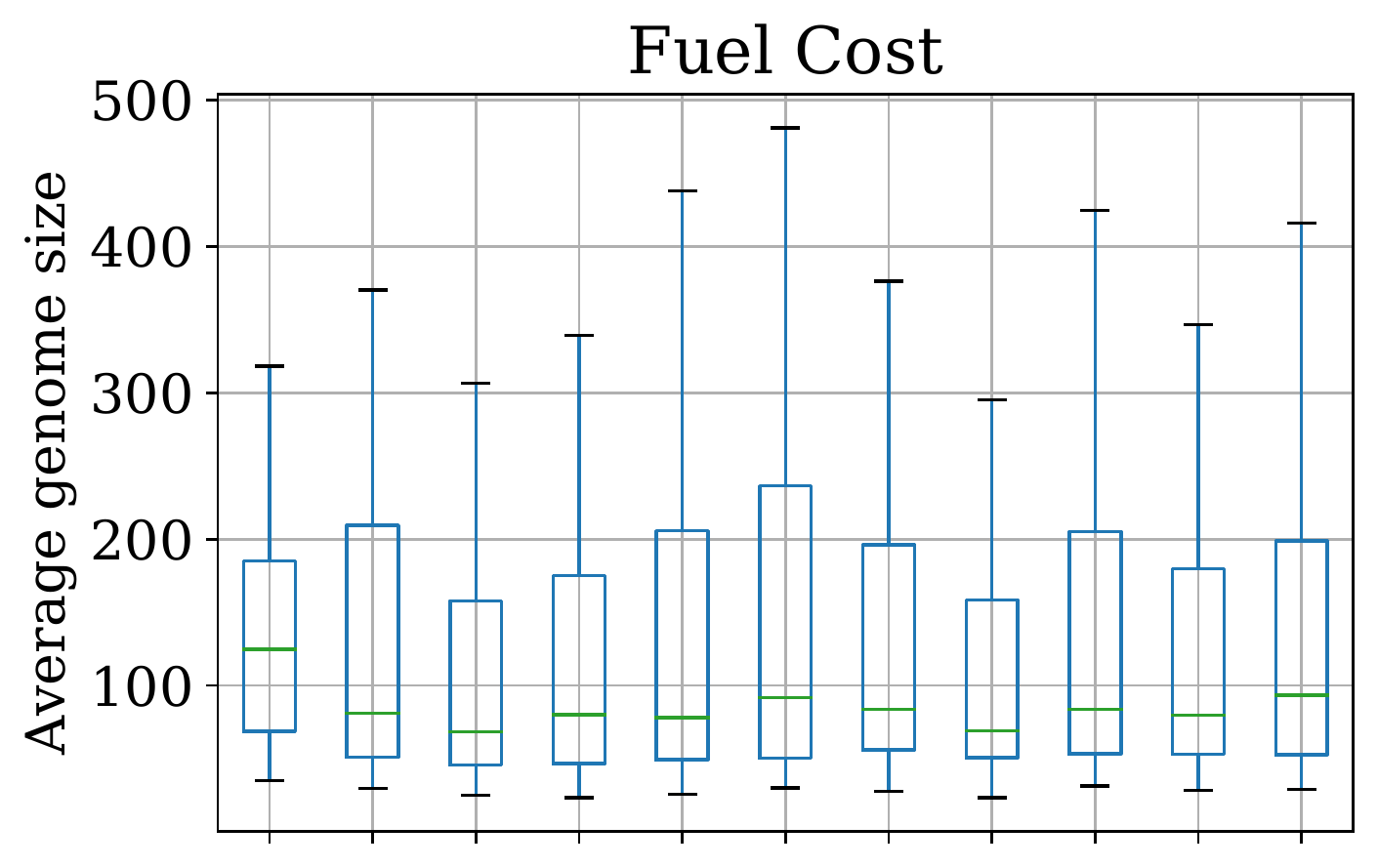}
    \label{fig:sub_push_size_3}
  \end{subfigure}
  \hfill
  \begin{subfigure}[b]{0.45\linewidth}
    \centering
    \includegraphics[width=\linewidth]{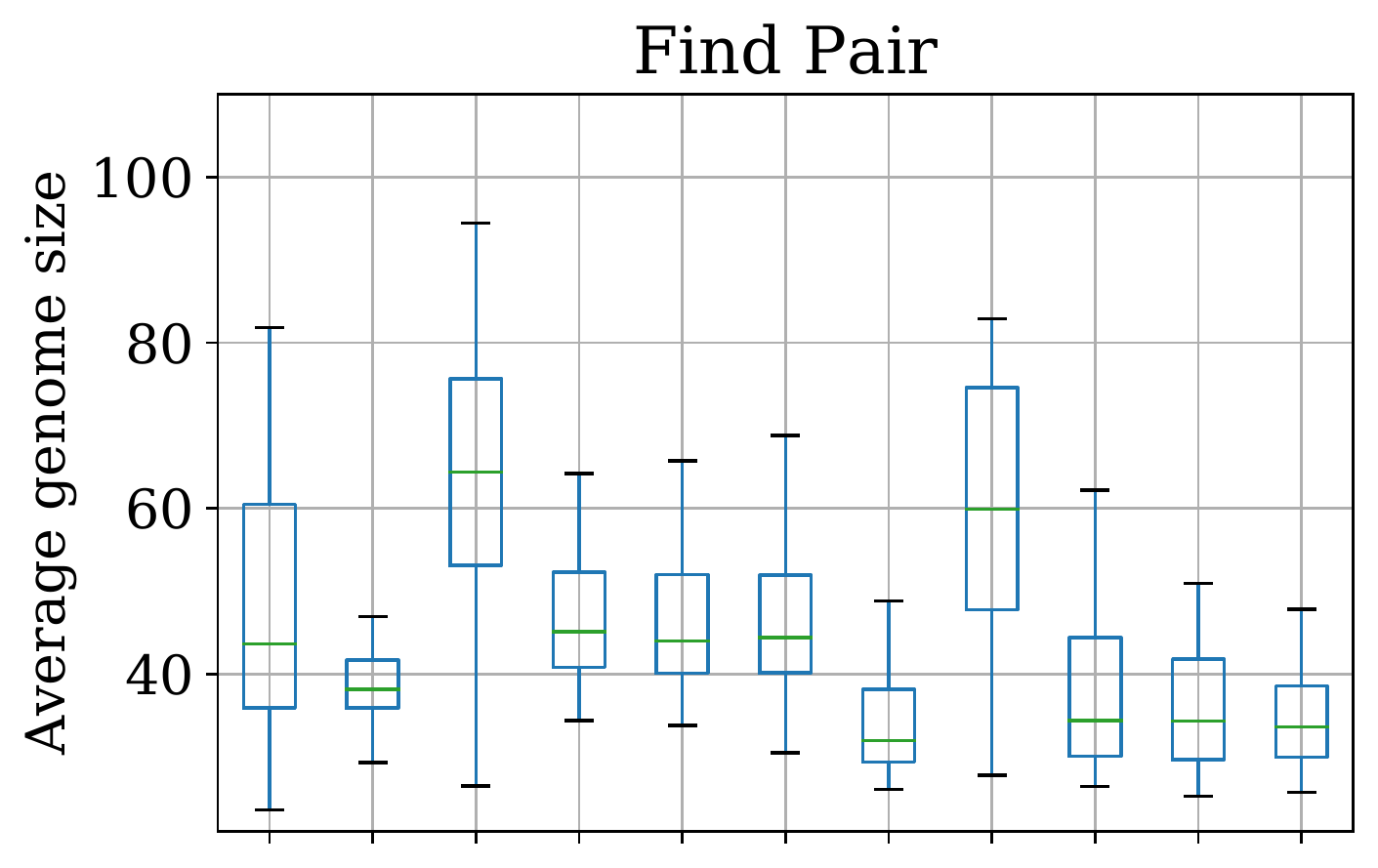}
    \label{fig:sub_push_size_4}
  \end{subfigure}

  \vspace{0.2cm}

  \begin{subfigure}[b]{0.45\linewidth}
    \centering
    \includegraphics[width=\linewidth]{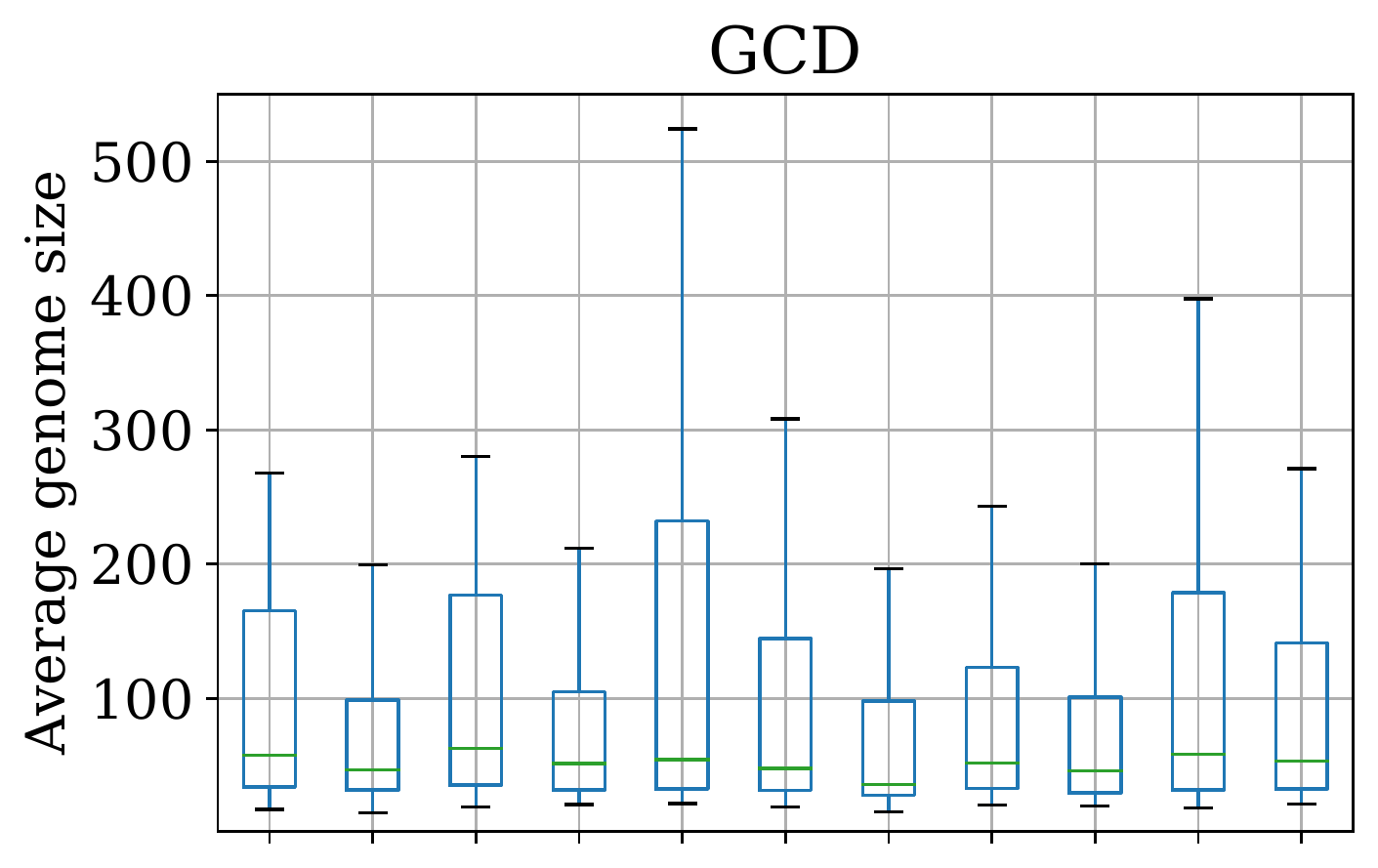}
    \label{fig:sub_push_size_5}
  \end{subfigure}
  \hfill
  \begin{subfigure}[b]{0.45\linewidth}
    \centering
    \includegraphics[width=\linewidth]{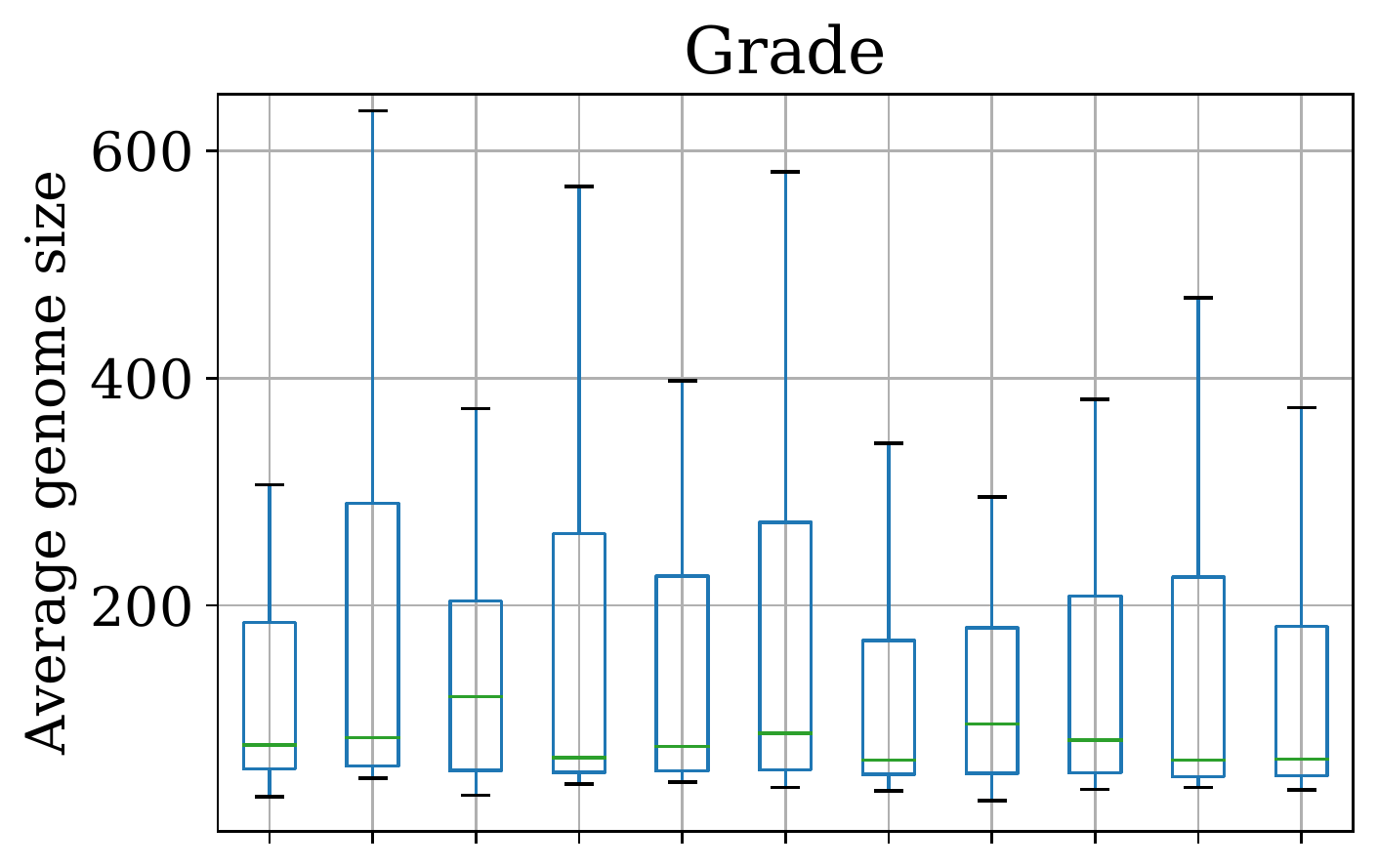}
    \label{fig:sub_push_size_6}
  \end{subfigure}
  
  \vspace{0.2cm}
  
  \begin{subfigure}[b]{0.45\linewidth}
    \centering
    \includegraphics[width=\linewidth]{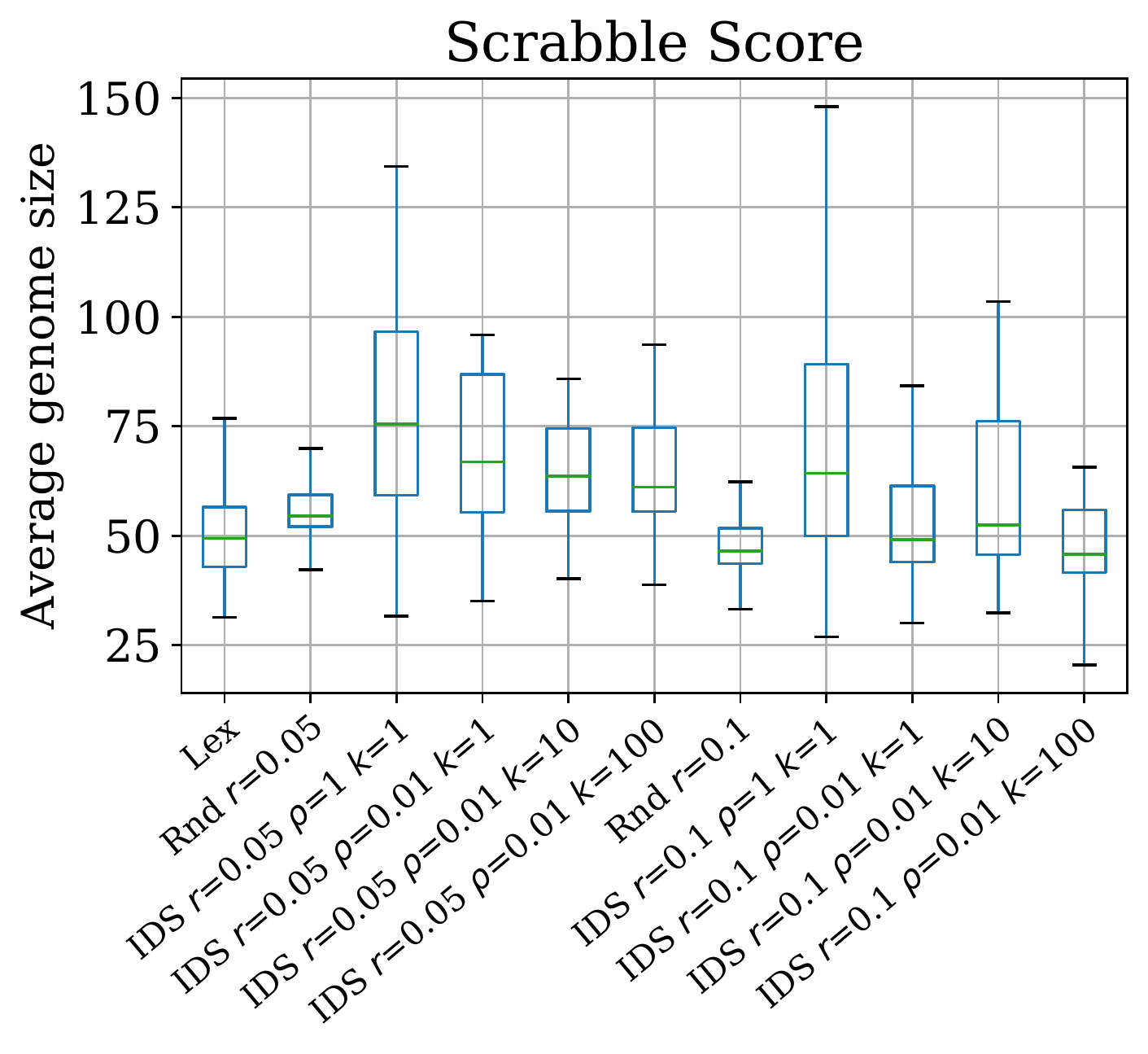}
    \label{fig:sub_push_size_7}
  \end{subfigure}
  \hfill
  \begin{subfigure}[b]{0.45\linewidth}
    \centering
    \includegraphics[width=\linewidth]{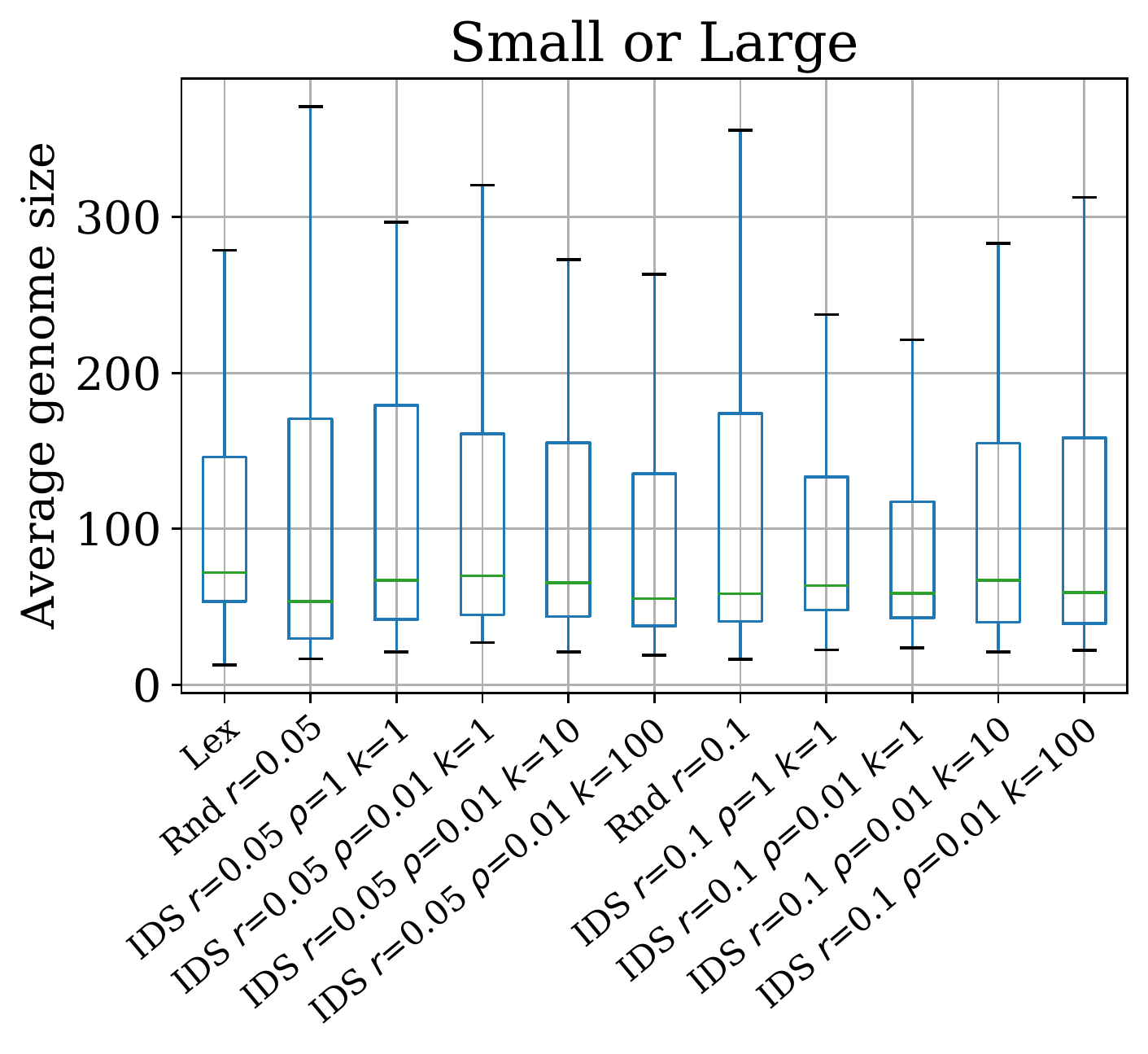}
    \label{fig:sub_push_size_8}
  \end{subfigure}
  
  \caption{Average genome length for the PushGP programs.}
  \label{fig:push_size}
\end{figure}

\begin{figure}
  \centering
  \begin{subfigure}[b]{0.45\linewidth}
    \centering
    \includegraphics[width=\linewidth]{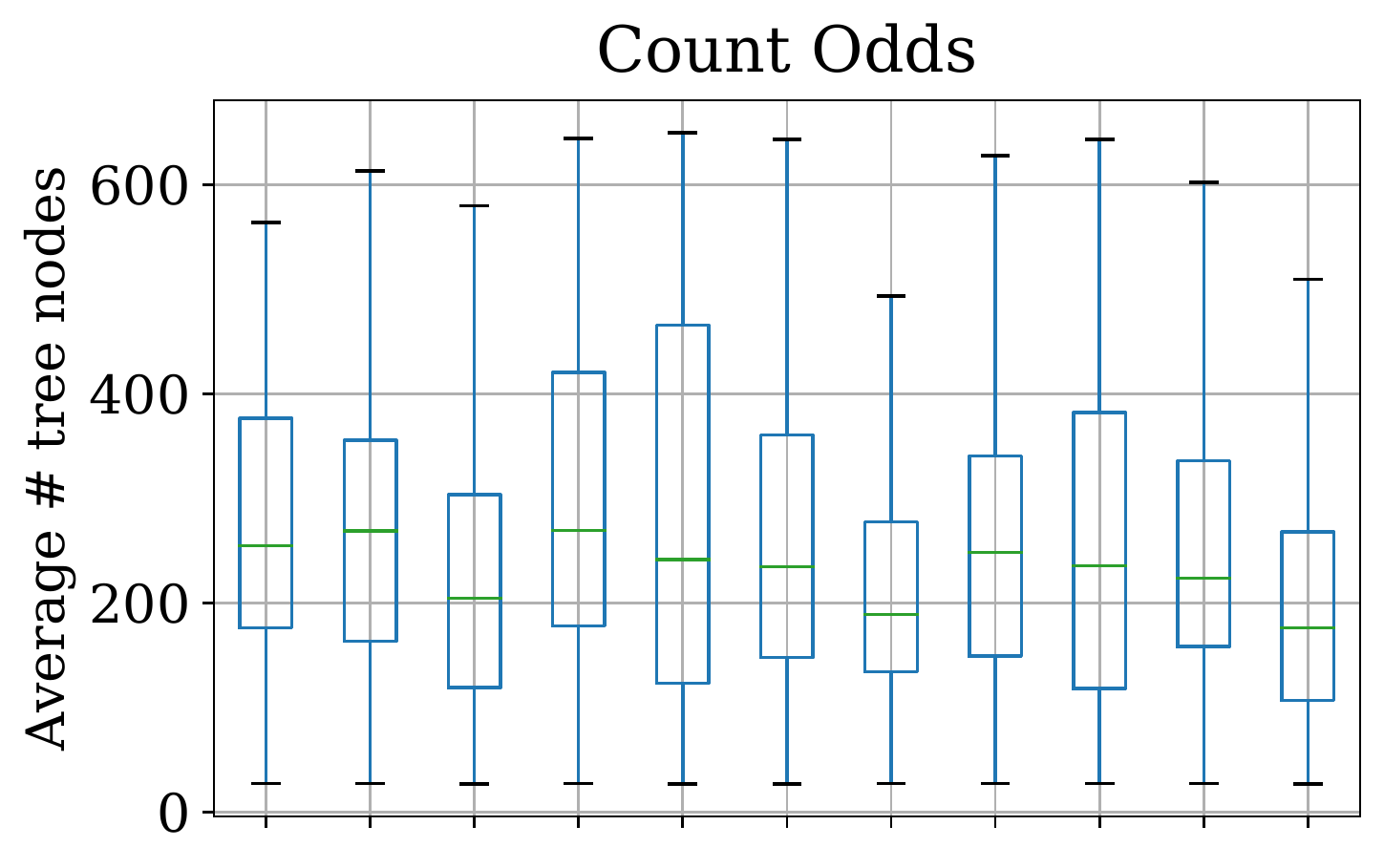}
    \label{fig:sub_g3p_size_1}
  \end{subfigure}
  \hfill
  \begin{subfigure}[b]{0.45\linewidth}
    \centering
    \includegraphics[width=\linewidth]{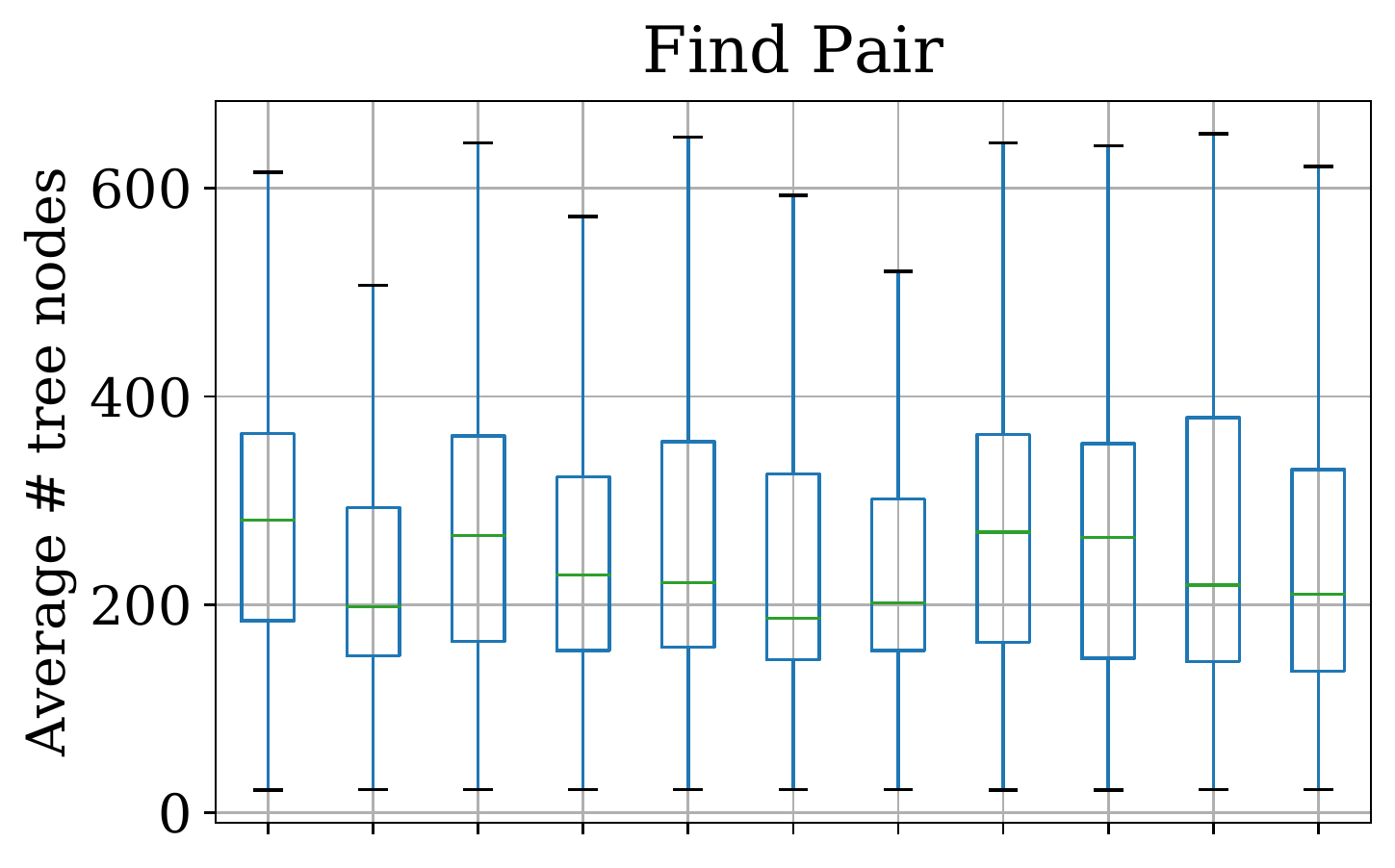}
    \label{fig:sub_g3p_size_2}
  \end{subfigure}
  
  \vspace{0.2cm}
  
  \begin{subfigure}[b]{0.45\linewidth}
    \centering
    \includegraphics[width=\linewidth]{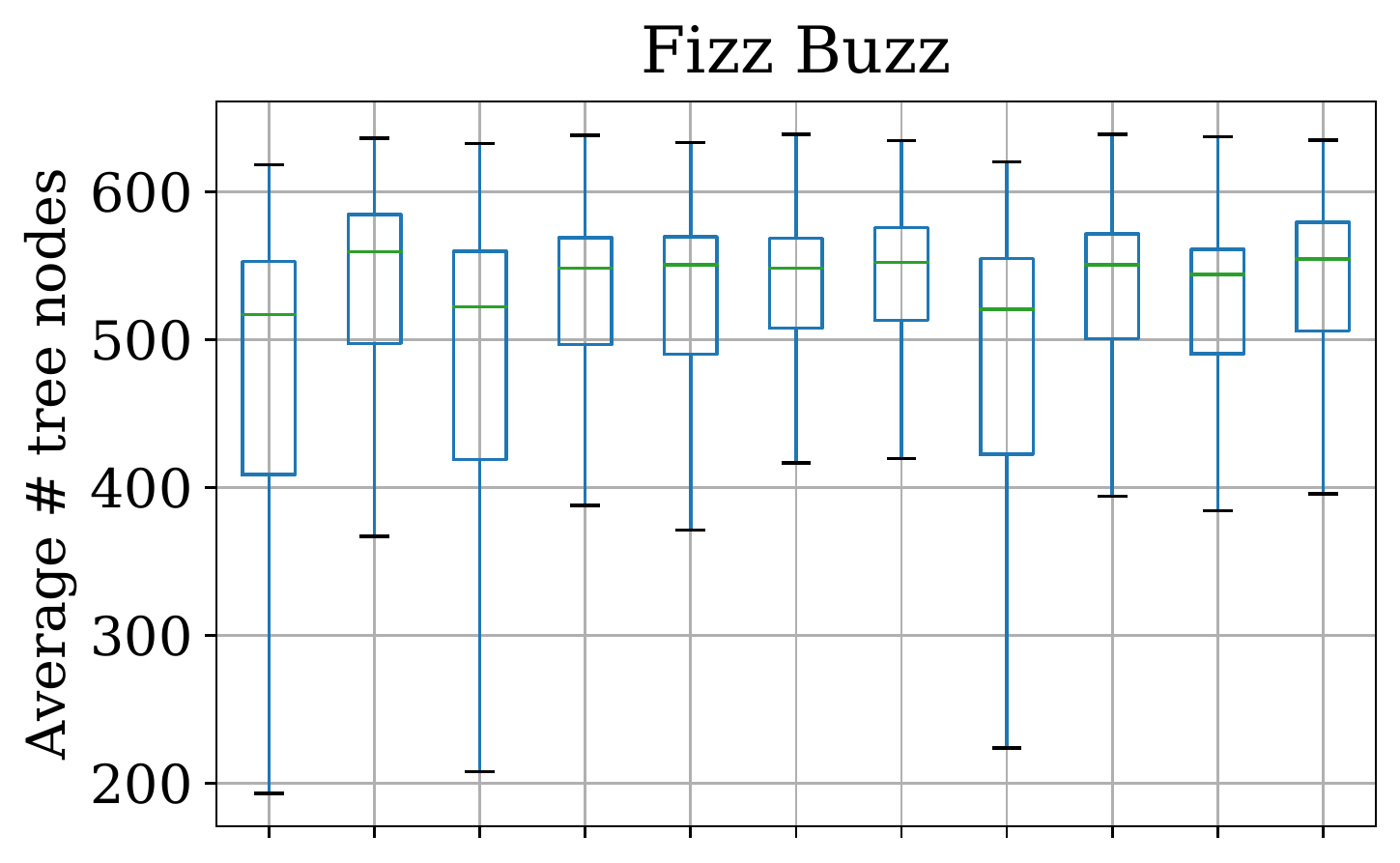}
    \label{fig:sub_g3p_size_3}
  \end{subfigure}
  \hfill
  \begin{subfigure}[b]{0.45\linewidth}
    \centering
    \includegraphics[width=\linewidth]{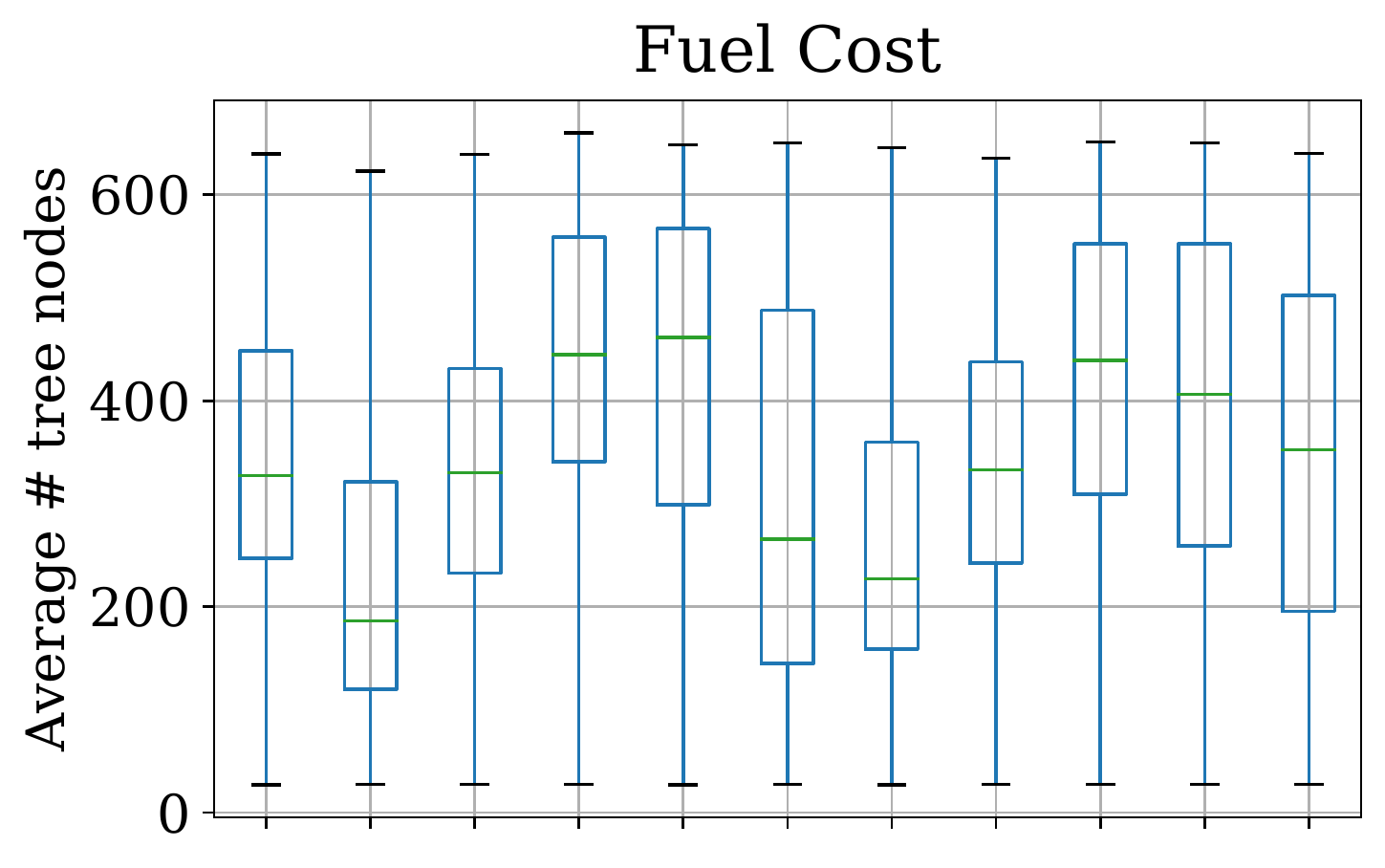}
    \label{fig:sub_g3p_size_4}
  \end{subfigure}

  \vspace{0.2cm}

  \begin{subfigure}[b]{0.45\linewidth}
    \centering
    \includegraphics[width=\linewidth]{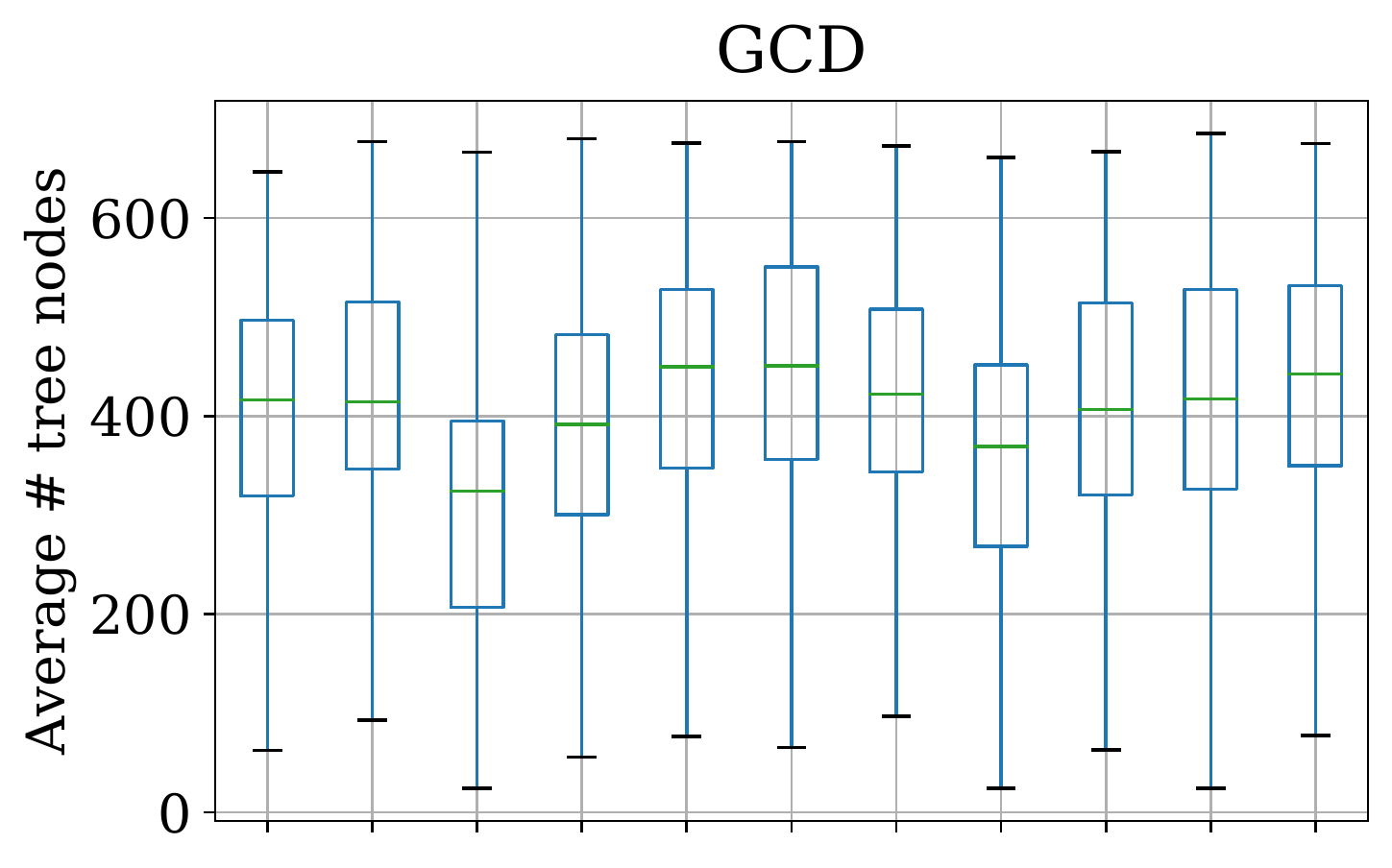}
    \label{fig:sub_g3p_size_5}
  \end{subfigure}
  \hfill
  \begin{subfigure}[b]{0.45\linewidth}
    \centering
    \includegraphics[width=\linewidth]{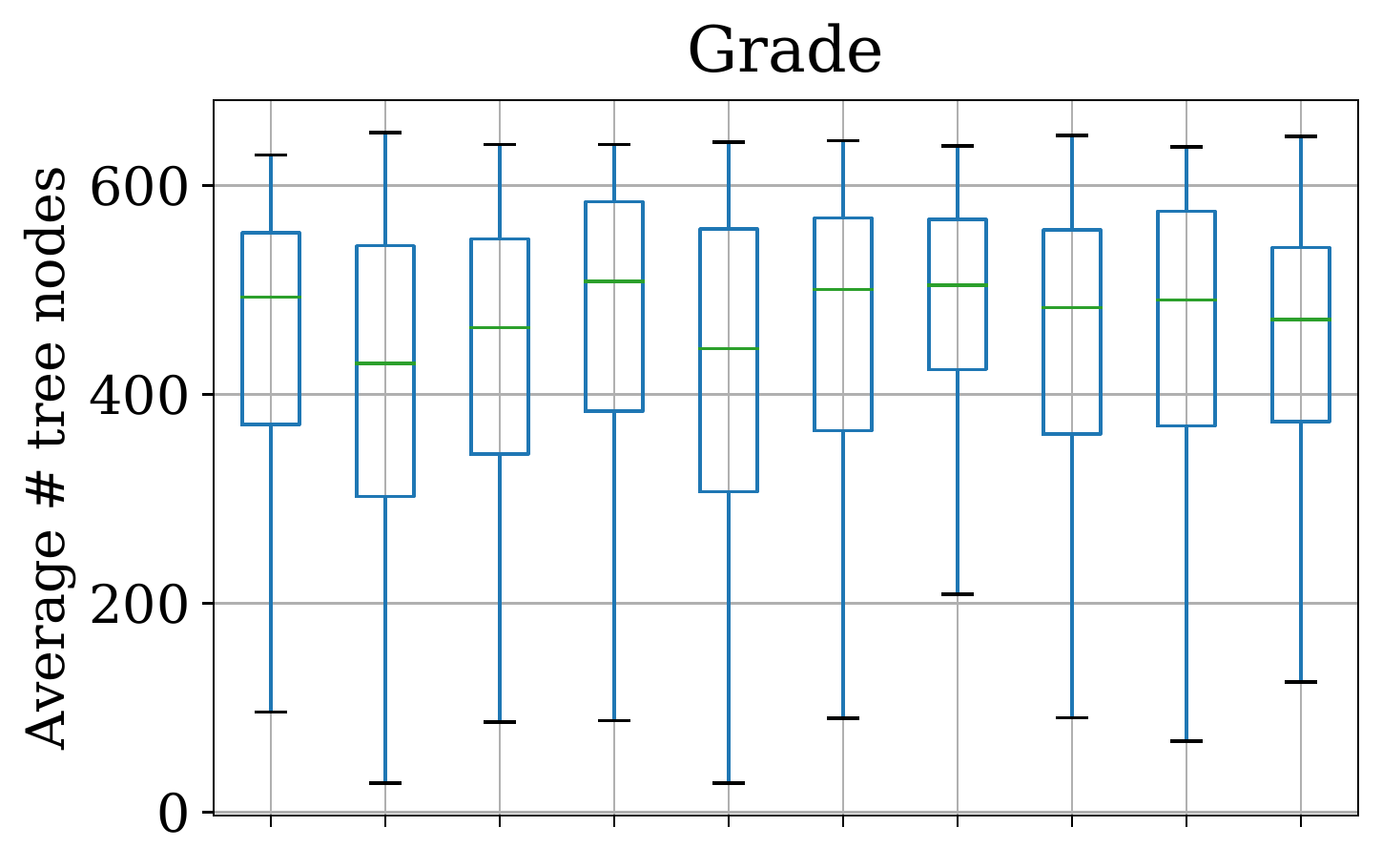}
    \label{fig:sub_g3p_size_6}
  \end{subfigure}
  
  \vspace{0.2cm}
  
  \begin{subfigure}[b]{0.45\linewidth}
    \centering
    \includegraphics[width=\linewidth]{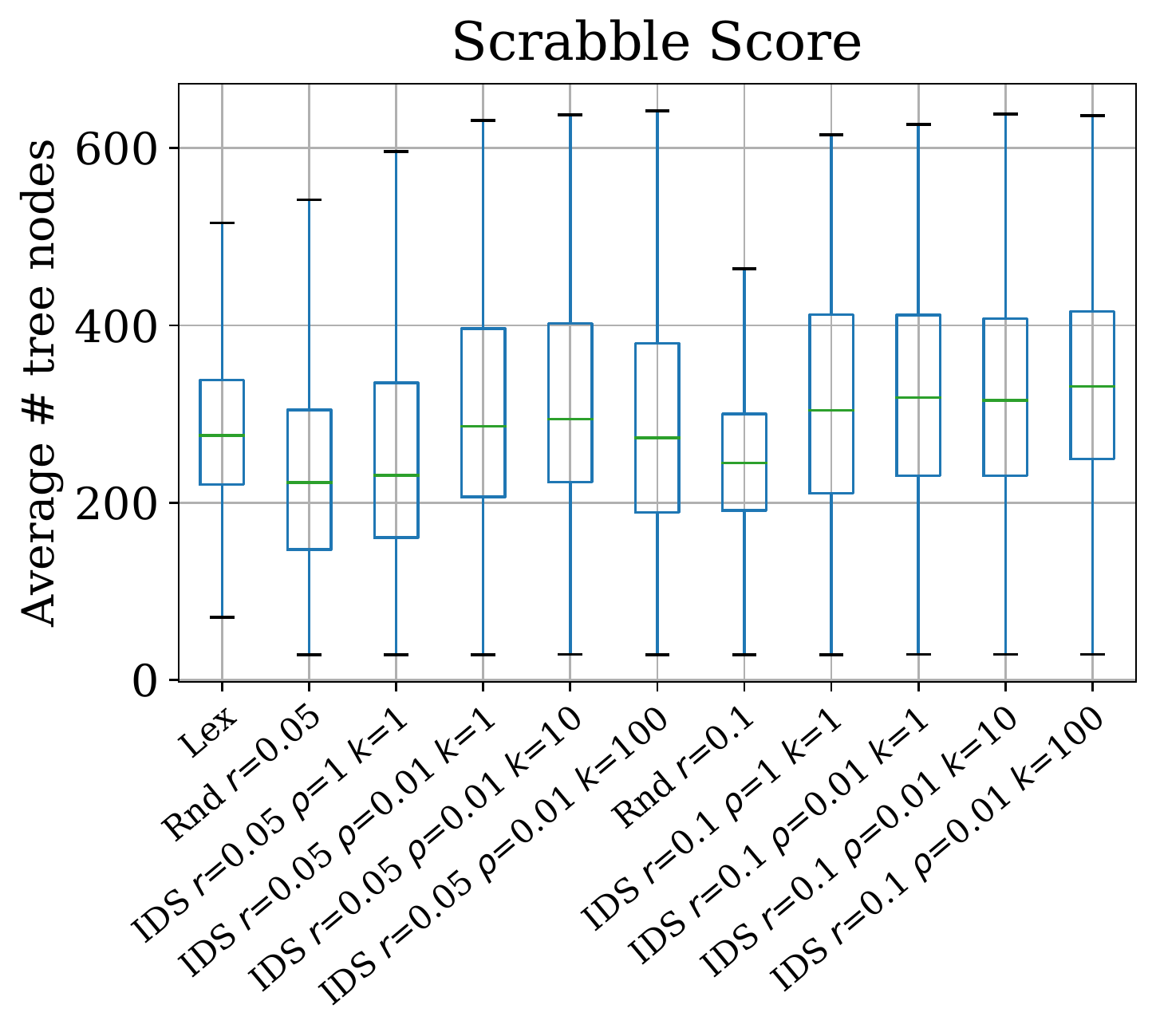}
    \label{fig:sub_g3p_size_7}
  \end{subfigure}
  \hfill
  \begin{subfigure}[b]{0.45\linewidth}
    \centering
    \includegraphics[width=\linewidth]{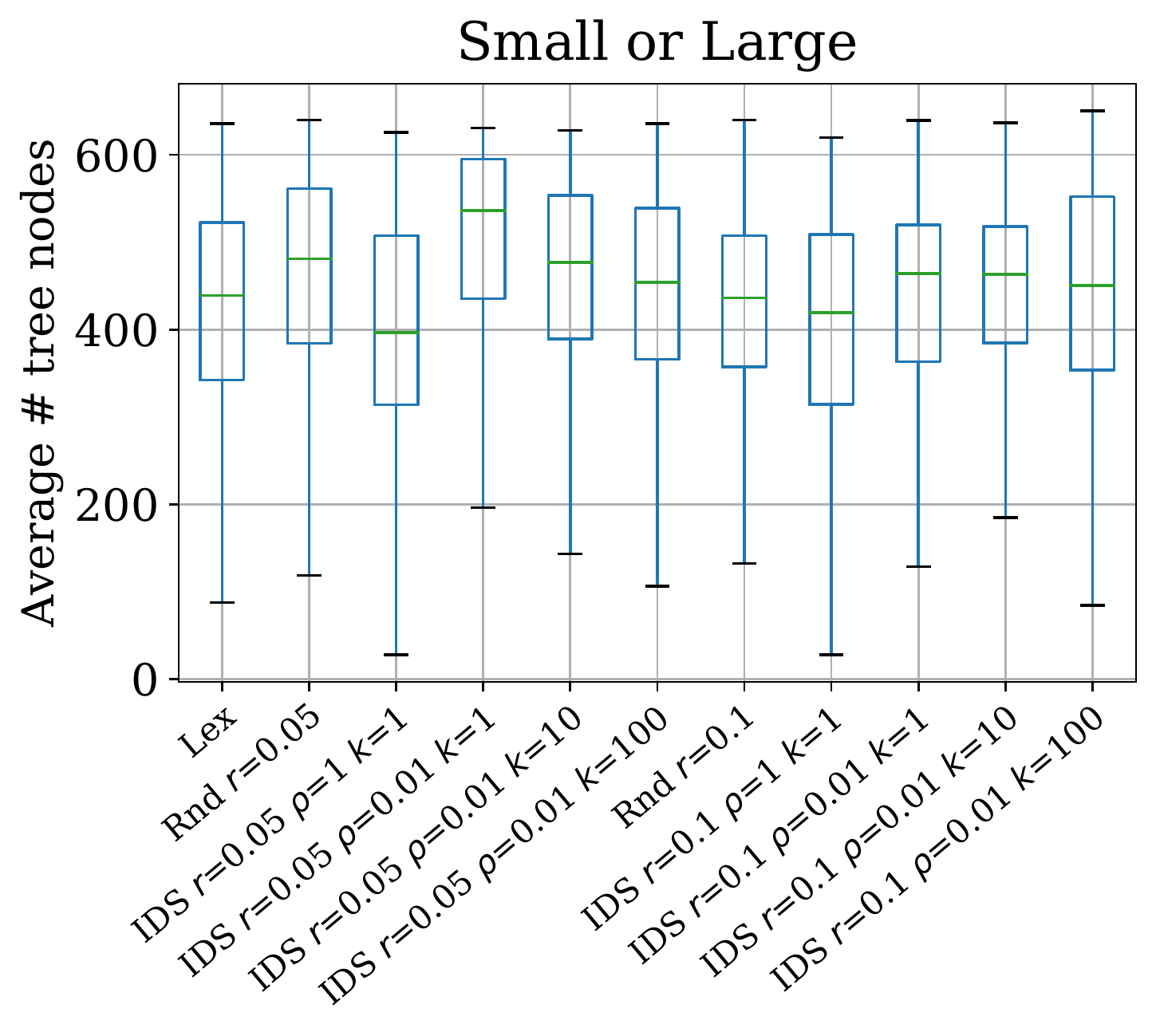}
    \label{fig:sub_g3p_size_8}
  \end{subfigure}
  
  \caption{Average number of tree nodes for the G3P programs.}
  \label{fig:g3p_size}
\end{figure}

Figure~\ref{fig:push_size} displays the distribution of the average genome length of all synthesized PushGP programs. This is reported across all runs for each problem and configuration combination. Similarly, figure~\ref{fig:g3p_size} displays the distribution of the average number of tree nodes of a generation of programs synthesized by G3P.

\section{Problem Descriptions}\label{appendix:problems}

\begin{table}[t]
\caption{Program synthesis benchmark problems selected from the first and second general program synthesis benchmark suite, along with their respective input and output types and multiplicities.}
\begin{tabular*}{\textwidth}{c @{\extracolsep{\fill}} lccc}
\hline
\textbf{Problem} & \textbf{Suite} & \textbf{Input Type } & \textbf{Output Type} \\  
\hline
Count Odds & PSB1 & Vector of Integer & Integer\\
Find Pair & PSB2 & Vector of Integer & Two Integers  \\
Fizz Buzz & PSB2 & Integer & String \\
Fuel Cost & PSB2 & Vector of Integer & Integer \\
GCD & PSB2 & Two Integers & Integer \\
Grade & PSB1 & Five Integers & String \\
Scrabble Score & PSB1 & String & Integer\\
Small or Large  & PSB1 & Integer & String \\
\hline
\end{tabular*}
\label{tab:problems}

\end{table}

These problem descriptions were modified from \cite{Helmuth2015psb1} and \cite{helmuth2021psb2}. The modifications are indicated with an \underline{underline}.
\begin{itemize}
    \item Count Odds - Given a vector of integers, 
return the number of integers that are odd, without
use of a specific \texttt{even} or \texttt{odd} instruction (but allowing instructions such as \texttt{mod} and \texttt{quotient}).
    \item Find Pair - Given a vector of integers, return the two  elements that sum to a target integer. 
    \item Fizz Buzz - Given an integer $x$, return ``Fizz" if $x$ is divisible by 3, "Buzz" if $x$ is divisible by $5$, "FizzBuzz" if $x$ is divisible by 3 and 5, and a string version of $x$ if none of the above hold.
    \item Fuel Cost - Given a vector of positive integers, divide each by $3$, round the result down to the nearest integer, and subtract $2$. Return the sum of all of the new integers in the vector.
    \item GCD - Given two integers, return the largest integer that divides each of the integers evenly.
    \item Grade - Given 5 integers, the first four represent the lower numeric thresholds for achieving an A, B, C, and D, and will be distinct and in descending order. The fifth represents the student’s numeric grade. The program must print \underline{one of A, B, C, D, or F as the achieved grade} depending on the thresholds and the numeric grade.
    \item Scrabble Score - Given a string of visible ASCII characters, return the Scrabble score for that string. Each letter has a corresponding value according to normal Scrabble rules, and non-letter characters are worth zero.

    \item Small or Large - Given an integer $n$, print "small" if $n < 1000$ and "large" if $n \geq 2000$ (and nothing if $1000 \leq n < 2000$).
\end{itemize}

\section{Example of a Push Instruction Set and G3P Grammar}\label{appendix:grammars}
Below is an example grammar (G3P) and instruction set (PushGP) used for the fizz buzz problem with both systems to illustrate their relative equivalency. A full list can be found in our web supplement \citep{boldi2023SupplementIDS}.

\begin{figure}
\begin{lstlisting}[frame=lines,numbers=none,xleftmargin=2.4em,framexleftmargin=2.4em,basicstyle=\small]
    
boolean     :boolean_and :boolean_deep_dup :boolean_dup
            :boolean_dup_items :boolean_dup_times 
            :boolean_empty :boolean_eq :boolean_flush 
            :boolean_from_integer :boolean_invert_first_then_and 
            :boolean_invert_second_then_and :boolean_not
            :boolean_or :boolean_pop :boolean_print :boolean_rot
            :boolean_shove :boolean_stack_depth :boolean_swap 
            :boolean_xor :boolean_yank :boolean_yank_dup 
            
exec        :exec_deep_dup :exec_do_count :exec_do_range :exec_do_times 
            :exec_do_while :exec_dup :exec_dup_items :exec_dup_times 
            :exec_empty :exec_eq :exec_flush :exec_if :exec_k :exec_pop 
            :exec_print :exec_rot :exec_s :exec_shove :exec_stack_depth 
            :exec_swap :exec_when :exec_while :exec_y :exec_yank 
            :exec_yank_dup 

integer     :integer_add :integer_dec :integer_deep_dup
            :integer_dup :integer_dup_items :integer_dup_times
            :integer_empty :integer_eq :integer_flush 
            :integer_from_boolean :integer_from_string 
            :integer_gt :integer_gte :integer_inc :integer_lt 
            :integer_lte :integer_max :integer_min :integer_mod 
            :integer_mult :integer_pop :integer_print :integer_quot
            :integer_rot :integer_shove :integer_stack_depth
            :integer_subtract :integer_swap :integer_yank
            :integer_yank_dup 

input       :in1

print       :print_newline 

string      :string_butlast :string_concat :string_contains 
            :string_deep_dup :string_drop :string_dup :string_dup_items 
            :string_dup_times :string_empty :string_empty_string
            :string_eq :string_flush :string_from_boolean
            :string_from_integer :string_length :string_parse_to_chars 
            :string_pop :string_print :string_replace :string_replace_first
            :string_rest :string_reverse :string_rot :string_shove 
            :string_split :string_stack_depth :string_substr :string_swap 
            :string_take :string_yank :string_yank_dup

constants   "Fizz" "Buzz" "FizzBuzz" 0 3 5


\end{lstlisting}
\caption{Instructions used for the Fizz Buzz problem with PushGP}
\end{figure}

\begin{figure}[!ht]
\centering
\begin{lstlisting}[frame=lines,numbers=none,xleftmargin=0em,framexleftmargin=0em,basicstyle=\small]
<predefined>              ::= 'i0 = int(); i1 = int(); i2 = int()'
                              'b0 = bool(); b1 = bool(); b2 = bool()'
                              's0 = str(); s1 = str(); s2 = str()'
                              'res0 = str()'
<code>
<code>                    ::= <code><statement>'\n'|<statement>'\n'
<statement>               ::= <assign>|<if>
<assign>                  ::= <bool_var>' = '<bool>|<int_assign>|
                              <string_var>' = '<string>|
                              <string_var>' = str('<int_var>')'
<number>                  ::= <number><num>|<num>
<num>                     ::= '0'|'3'|'5'
<comp_op>                 ::= '<'|'>'|'=='|'>='|'<='|'!='
<bool_var>                ::= 'b0'|'b1'|'b2'
<bool>                    ::= <bool_var>|<bool_const>|'not '<bool>|
                              '( '<bool>' '<bool_op>' '<bool>' )'|
                              <int>' '<comp_op>' '<int>|<string>' in 
                              '<string>|<string>' not in '<string>|
                              <string>' == '<string>|<string>' !=
                              '<string>|
                              <string>'.startswith('<string>')'|
                              <string>'.endswith('<string>')'
<bool_op>                 ::= 'and'|'or'
<bool_const>              ::= 'True'|'False'
<if>                      ::= 'if '<bool>':{:\n'<code>':}'|
                              'if '<bool>':{:\n'<code>':}else:{:
                              \n'<code>':}'
<int_var>                 ::= 'i0'|'i1'|'i2'|'in0'
<int_assign>              ::= <int_var>' = '<int>|
                              <int_var>' '<arith_ops>'= '<int>
<int>                     ::= <int_var>|'int('<number>'.0)'|
                              <arith_prefix><int>|
                              '( '<int>' '<arith_ops>' '<int>' )'|
                              <int_arith_ops_protected>'('<int>',
                              '<int>')'|'min('<int>', '<int>')'|
                              'max('<int>', '<int>')'|'abs('<int>')'|
                              'len('<string>')'|'saveOrd('<string>')'
<arith_ops>               ::= '+'|'-'|'*'
<int_arith_ops_protected> ::= 'divInt'|'mod'
<arith_prefix>            ::= '+'|'-'
<string_var>              ::= 's0'|'s1'|'s2'|'res0'
<string>                  ::= <string_var>|<string_const>|
                              <string_slice>|
                              'getCharFromString('<string>', '<int>')'|
                              'saveChr('<int>')'|'('<string>' +
                              '<string>')'|<string>'.lstrip()'|
                              <string>'.rstrip()'|<string>'.strip()'|
                              <string>'.lstrip('<string>')'|
                              <string>'.rstrip('<string>')'|
                              <string>'.strip('<string>')'|
                              <string>'.capitalize()'
<string_slice>            ::= <string>'['<int>':'<int>']'|
                              <string>'[:'<int>']'|<string>'['<int>':]'
<string_const>            ::= "'Fizz'"|"'Buzz'"|"'FizzBuzz'"
\end{lstlisting}
\caption{Grammar used for the Fizz Buzz problem with G3P}
\label{fig:progsys_grammar}
\end{figure}

\end{document}